\documentclass[times, review, 10pt]{elsarticle}

\usepackage{xcolor}
\usepackage{multirow}
\usepackage{makecell}
\usepackage{stmaryrd}
\usepackage{booktabs}

\usepackage{amssymb}
\usepackage{amsmath}
\usepackage{setspace}
\usepackage{array}
\usepackage{tabularx}
\usepackage{calc}
\usepackage{subfig}
\journal{  }
\usepackage[dvipsnames]{xcolor}

\usepackage{amsmath,amsfonts,bm}

\usepackage{bbm}

\usepackage{nicefrac}       
\usepackage{microtype}      
\usepackage{color}
\usepackage{array}
\usepackage{multirow}
\usepackage{dirtytalk}

\usepackage{amsthm}
\usepackage[para,online,flushleft]{threeparttable}
\def\argmin{\mathop{\mathrm{arg}\, \mathrm{min}}\limits}

\newcommand{\comment}[1]{}
\begin{document}

\begin{frontmatter}



\title{Robust Multi-Modal Face Anti-Spoofing with Domain Adaptation: Tackling Missing Modalities, Noisy Pseudo-Labels, and Model Degradation}



\author[2,fn1]{Ming-Tsung Hsu}
\ead{xmc510063@gapp.nthu.edu.tw}

\author[2]{Fang-Yu Hsu}
\ead{shellyhsu@gapp.nthu.edu.tw}

\author[2]{Yi-Ting Lin}
\ead{yitinglin@gapp.nthu.edu.tw}

\author[2]{Kai-Heng Chien}
\ead{ss113062582@gapp.nthu.edu.tw}

\author[2]{Jun-Ren Chen}
\ead{jr.chen.088@gapp.nthu.edu.tw}

\author[2]{Cheng-Hsiang Su}
\ead{vincent071659@gapp.nthu.edu.tw}

\author[2]{Yi-Chen Ou}
\ead{easonou1105@gapp.nthu.edu.tw}

\author[2]{Chiou-Ting Hsu}
\ead{cthsu@cs.nthu.edu.tw}

\author[1,2]{Pei-Kai Huang\corref{cor1}}
\ead{alwayswithme@fjnu.edu.cn}

\cortext[cor1]{Corresponding author}
\fntext[fn1]{This is the first author.}

\affiliation[1]{organization={College of Computer and Cyber Security,  Fujian Normal University},
                city={Fuzhou},
                country={China}
                }

\affiliation[2]{organization={Department of Computer Science, 
                              National Tsing Hua University},
                city={Hsinchu},
                country={Taiwan}}

\begin{abstract}
Recent multi-modal face anti-spoofing (FAS) methods have investigated the potential of leveraging multiple modalities to distinguish live and spoof faces.
However, pre-adapted multi-modal FAS models often fail to  detect unseen attacks from new target domains.
Although a more realistic domain adaptation (DA) scenario has been proposed for single-modal FAS to learn specific spoof attacks during inference, DA remains unexplored in multi-modal FAS methods.
In this paper, we propose a novel framework, MFAS-DANet, to address three major challenges in multi-modal FAS under the DA scenario: missing modalities, noisy pseudo labels, and model degradation.
First, to tackle the issue of missing modalities, we propose extracting complementary features from other modalities to substitute missing modality features or enhance existing ones.
Next, to reduce the impact of noisy pseudo labels during model adaptation, we propose deriving reliable pseudo labels by leveraging prediction uncertainty across different modalities.  
Finally, to prevent model degradation, we design an adaptive mechanism that decreases the loss weight during unstable adaptations and increasing it during stable ones. 
Extensive experiments demonstrate the effectiveness and state-of-the-art performance of our proposed MFAS-DANet. 
\end{abstract}




\end{frontmatter}

\section{Introduction}
 
\label{sec:introduction}

With the rapid advancement of facial recognition technology, many applications, such as biometric authentication, access control, and payment systems, have come to rely on face recognition to enhance the convenience of everyday activities.
However, these applications face increasingly serious challenges from the evolving threat of facial spoofing attacks, which pose significant security risks.
As facial recognition applications encounter increasingly sophisticated facial spoofing attacks, a variety of face anti-spoofing (FAS) techniques \cite{huang2024survey} have been developed by utilizing either single (RGB) or multiple (RGB, infrared, depth) modalities to counter this evolving threat. 

\begin{figure}[t]
\centering
\includegraphics[width=1 \linewidth]{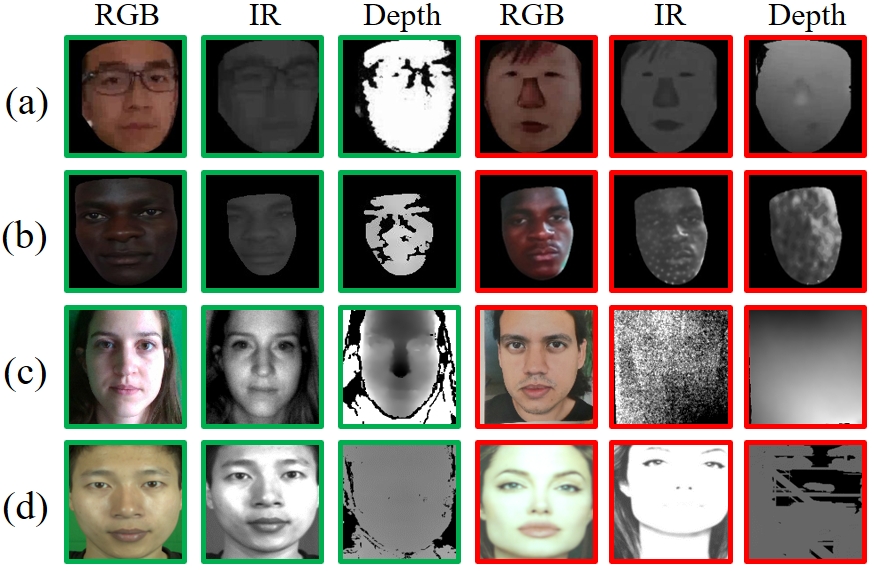} 
\caption{Examples of live faces (boxes in green) and spoof faces (boxes in red).
From these samples, we observe that different multi-modal FAS datasets exhibit significant domain gaps.
}
\label{fig:data examples}
\end{figure}

\begin{figure}[t]  
    \centering
    \includegraphics[width=1 \textwidth]{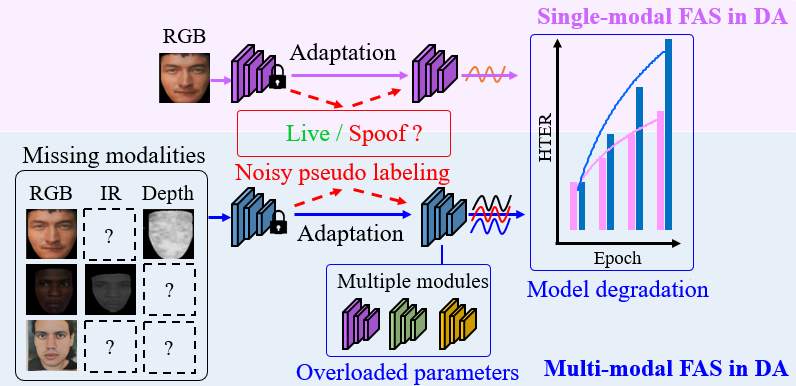}  
\caption{   
Challenges in single-modal and multi-modal face anti-spoofing (FAS) during domain adaptation (DA). Noisy pseudo labels often cause model degradation in both methods, while multi-modal FAS encounters additional issues, including missing modalities and overloaded parameters, which may potentially accelerate this degradation.  
} 
    \label{fig:idea}  
\end{figure}

Most single-modal / multi-modal FAS methods adopt domain generalization (DG) scenario to counter newly developed facial spoof attacks from unseen domains.
Since RGB images exhibit rich texture and color information, single-modal FAS methods \cite{jia2020single, liao2023domain, huang2022learnable, huang2023ldcformer} focus on learning discriminative liveness features to enhance the generalization ability of FAS models.
To address newly developed facial spoof attacks from unseen domains, most single-modal \cite{jia2020single, liao2023domain, huang2022learnable, huang2023ldcformer} and multi-modal FAS methods \cite{yu2020multi, lin2024suppress, liu2023fm, liu2023ma} adopt a DG  scenario to enhance model generalization. 
In addition, as noted in \cite{lin2024suppress}, different modalities are more effective for addressing specific attack types under different deployment conditions.
For example, IR modalities perform well in low-light conditions, while depth modalities are effective in capturing 3D shape information to detect replay attacks.
Therefore, multi-modal FAS methods \cite{yu2020multi, lin2024suppress, liu2023fm, liu2023ma}, which incorporate modalities (such as infrared and depth maps) beyond RGB, can significantly enhance the detection performance compared to single modal FAS methods.
Nevertheless, 
as observed in Figures~\ref{fig:idea} and \ref{fig:data examples}, the same modalities captured by different sensors may exhibit significant domain shifts. 
These offline pre-adapted FAS models remain ineffective at detecting newly developed spoofing attacks in a new target domain \cite{Rostami_2021_ICCV, george2019biometric}. Compared to DG scenario, domain adaptation (DA) scenario uses available unlabeled target domain data to adapt source knowledge to the target domain during model adaptation.
While the exploration of DA has been studied in single-modal FAS task \cite{wang2021self}, its potential in multi-modal FAS remains unexplored.

Compared to single-modal FAS within DA, multi-modal FAS encounters additional challenges in this setting. First, as illustrated in Figure~\ref{fig:idea}, multi-modal FAS models may encounter target data with missing infrared (IR) or depth modalities during adaptation \cite{yu2023flexible}. Consequently, developing strategies to handle the absence of these modalities is crucial to enable accurate predictions in such scenarios.
The second challenge is noisy pseudo-labeling. Similar to single-modal FAS, multi-modal FAS encounters difficulty in determining reliable pseudo-labels for unseen target data due to the visual similarities between live and spoof faces, as shown in Figure~\ref{fig:idea}.
In addition, as noted in \cite{lin2024suppress}, the inherent low quality of modalities like depth and IR often yields unreliable multi-modal features and further exacerbates the noisy pseudo-labeling issue during adaptation.
Finally, the third challenge involves model degradation. 
The larger number of parameters in multi-modal FAS models complicates the adaptation process and renders stable adaptation more challenging than in single-modal models.
To sum up, multi-modal FAS in DA is additionally challenged by missing modalities, exacerbated noisy pseudo-labeling, and the issue of overloaded parameter, all of which may lead to increased model degradation compared to single-modal FAS.

In this paper, we address the aforementioned challenges in multi-modal FAS under DA scenario and propose a novel DA framework, MFAS-DANet, incorporating three key strategies: cross-modal feature transformation,  prediction-reliability aware pseudo-labeling, and model-stability aware adaptation. 
First, to address the missing modalities issue,
we propose an effective cross-modal feature transformation to extract complementary features from available modalities to substitute missing modality features or enhance existing ones.
Second, to mitigate noisy pseudo-labels, we propose a novel prediction-reliability aware pseudo-labeling that leverages prediction uncertainty across modalities to generate reliable pseudo-labels for effective model adaptation.
Finally, to address model degradation, we propose a model-stability aware adaptation that enables FAS models to stably learn new attack types and unseen domain information while mitigating degradation risks during the adaptation process at the inference stage.

Our contributions are summarized as follows: 
\begin{itemize} 
\item 
We propose a novel and practical framework, MFAS-DANet for multi-modal face anti-spoofing under the domain adaptation (DA) scenario. To the best of our knowledge, this is the first work  addressing the challenges of domain adaptation for multi-modal FAS. 
\item  
We propose three key strategies: cross-modal feature transformation, prediction-reliability aware pseudo-labeling, and model-stability aware adaptation, to tackle the issues of missing modalities, noisy pseudo-labeling, and model degradation in DA for multi-modal FAS.  
\item  
Extensive experiments demonstrate that the proposed MFAS-DANet surpasses previous DG techniques for the multi-modal FAS task and achieves state-of-the-art performance. 
\end{itemize}

\section{Related work}
\label{sec:related_work}

\subsection{Single-modal face anti-spoofing}

Previous single-modal FAS methods have leveraged Domain Generalization (DG) techniques to achieve impressive performance in handling unseen spoofing attacks.
Many studies have explored the use of auxiliary supervision to enhance the distinction between live and spoof faces. Depth maps\cite{huang2021face,liu2018learning,shao2019multi,yu2020face,wang2021self,zhou2023instance} and rPPG signals\cite{liu2018remote,liu2022learning,huang2021face} have been widely adopted as additional cues to improve model robustness against spoofing attacks.

More recently, to address the limitations of convolution-based architectures in modeling long-range dependencies, transformer-based FAS methods \cite{liao2023domain, huang2022adaptive, huang2023ldcformer, huang2025channel} have gained attention. These models leverage self-attention mechanisms to capture global contextual relationships, demonstrating superior adaptability compared to conventional CNN-based approaches.
In \cite{huang2025slip,huang2024one}, the authors proposed using handcrafted spoof cue maps to facilitate one-class face anti-spoofing.
In addition, Domain Adaptation (DA) techniques focus on 
adapting a model trained on source data to better fit the unlabeled target domain.
Recent work \cite{he2024category} has proposed aligning gradient vectors between source and target domains to eliminate gradient discrepancies. In addition, in \cite{mao2024weighted}, the author utilized optimal transport to align feature and pseudo-label distributions across domains to  improve the performance in cross-scenario tasks. In \cite{zhou2022generative}, target data is stylized to match the source domain, enhancing model compatibility. Moreover, \cite{yue2023cyclically} demonstrates that disentangling domain-invariant liveness features can further improve model robustness.

\subsection{Multi-modal face anti-spoofing}

To effectively counter the growing threat of spoof attacks,  recent  multi-modal face anti-spoofing methods methods \cite{george2021cross, yu2024rethinking, yu2024visual, lin2024suppress} have adopted domain generalization techniques to enhance the robustness of FAS models by using additional modalities, such as infrared (IR) and depth modalities.

In \cite{george2021cross}, the authors proposed a multi-head late fusion model with cross-modal focal loss to enhance RGB-D FAS by adjusting supervision based on channel confidence. Meanwhile, \cite{lin2024suppress} proposed cross-modal learning framework that mitigates unreliable modality influence and balances gradient optimization to improving the generalization of multi-modal FAS.
In \cite{liu2023fm, liu2023ma}, the authors proposed flexible, modality-agnostic architectures that adapt to different modalities for effective multi-modal fusion. The authors in \cite{antil2024mf2shrt} followed by optimizing fusion efficiency with a shared-layered transformer to reduce computational overhead.
The authors in \cite{zhang2021progressive} proposed a progressive modality cooperation framework for multi-modal domain adaptation, which facilitates feature alignment across modalities and domains.
While the exploration of domain adaptation has been studied in other computer vision tasks, the potential for leveraging domain adaptation to handle domain shifts of different modalities in multi-modal face anti-spoofing remains unexplored.


 \begin{figure}[t] 
    \centering 
    \includegraphics[width=13cm]{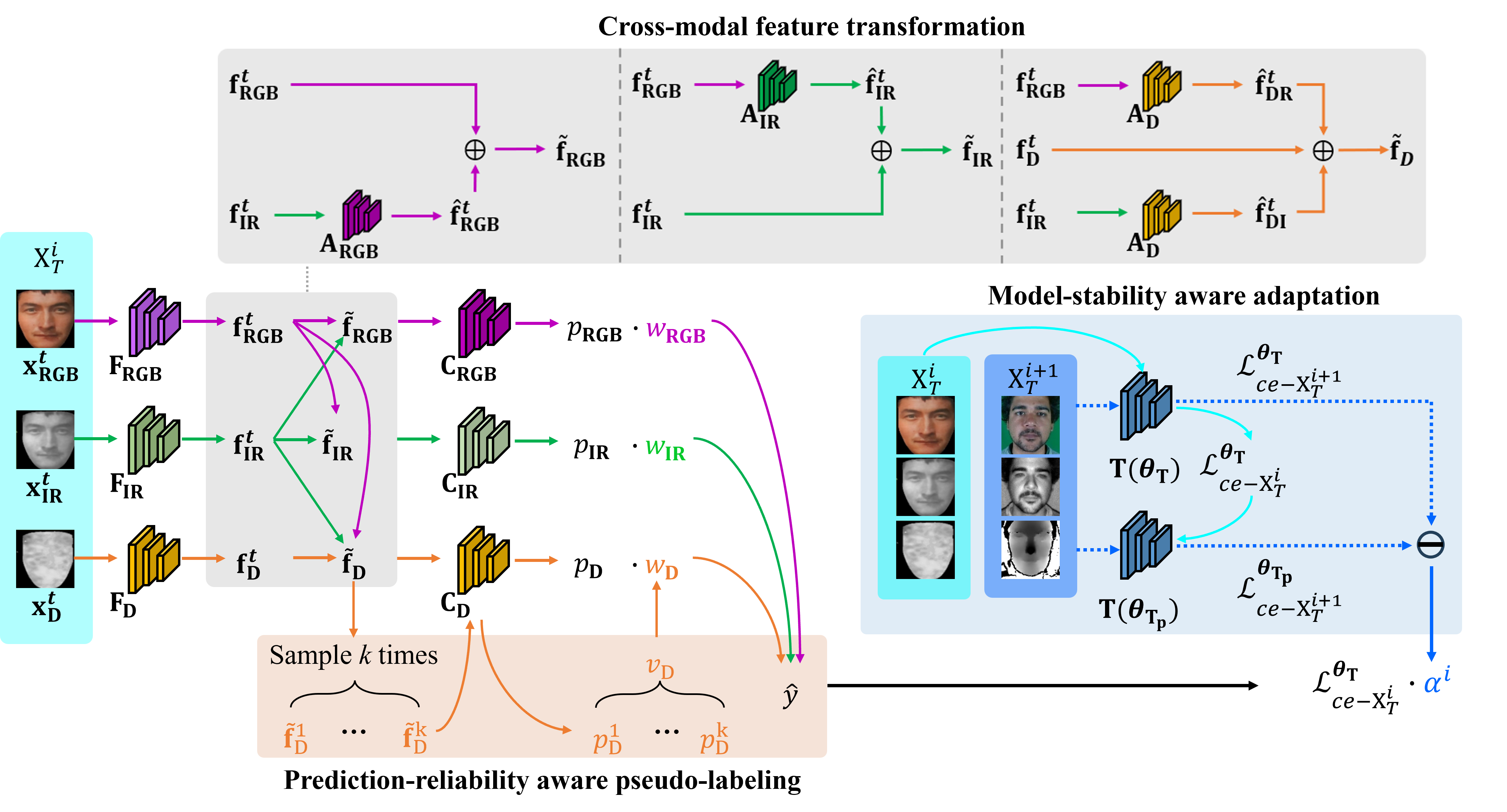}  
    \caption{Overview of the proposed MFAS-DANet, which includes three key strategies: cross-modal feature transformation, prediction-reliability aware pseudo-labeling, and model-stability aware adaptation, to address the challenges of missing modalities, noisy pseudo-labels, and model degradation.}     
\label{fig:Framework}  
\end{figure}

\section{Proposed method}
\label{sec:method}

In Section~\ref{sec:overview}, we first present the problem statement of domain adaptation in multi-modal face anti-spoofing  and give a brief overview of the proposed MFAS-DANet.
Next, in Sections~\ref{sec:feature enhancement}-\ref{sec:Model-reliability adaptation}, we present  three key strategies in the MFAS-DANet, including cross-modal feature transformation, prediction-reliability aware pseudo-labeling, and model-stability aware adaptation, respectively. 
   
\subsection{Problem Statement and Overview}
\label{sec:overview} 

In this paper, we address the problems of domain adaptation in multi-modal face anti-spoofing. 
Given the labeled source data 
$X_S = \left\{ \mathbf{x}_{\text{RGB}}^s, \right. \left. \mathbf{x}_{\text{IR}}^s,\, \mathbf{x}_{\text{D}}^s,\, y^s \right\}$,
we begin by training a multi-modal face anti-spoofing model $\mathbf{T}$ using the standard cross-entropy loss $\mathcal{L}_{\text{ce}}$, defined as: 
\begin{equation}
  \mathcal{L}_{ce}  =   \sum_{ \substack{\forall  \mathbf{f}_\text{mod}^s, \\ \text{mod} \in \{ \text{RGB}, \text{IR}, \text{D}\} }}  - y^s \log(\mathbf{C}_\text{mod}( \mathbf{f}^s_\text{mod}) ).
\label{eqn:entropy_source}
\end{equation}  

\noindent 
where $\mathbf{f}^s_\text{mod} = \mathbf{F}_\text{mod} (\mathbf{x}^s_\text{mod})$ denotes the liveness features of the source image $\textbf{x}^s_{\text{mod}}$, with $\text{mod} \in \{ \text{RGB},\text{IR},\text{D} \}$,
$\mathbf{F}_\text{mod}$ and $\mathbf{C}_\text{mod}$ denote the feature extractor and the classifier, respectively, and $y^s$ denotes the binary live/spoof labels. As RGB is the most readily accessible modality, we follow previous multi-modal FAS methods \cite{lin2024suppress} and specifically address scenarios where IR or depth modalities are missing during inference.
Hence, our goal is to adapt $\textbf{T}$ to the unlabeled target data $X_T$,  which may have missing modalities: $X_T = \left\{ \textbf{x}_{\text{RGB}}^t,\, \mathbb{I}(\textbf{x}_{\text{IR}}^t \neq \emptyset) \cdot \textbf{x}_{\text{IR}}^t, \right. \left. \mathbb{I}(\textbf{x}_{\text{D}}^t \neq \emptyset) \cdot \textbf{x}_{\text{D}}^t \right\}$, where $\mathbb{I}(\cdot)$ denotes the indicator function 
and $\textbf{x}_{\text{IR}}^t \neq \emptyset$ and $\textbf{x}_{\text{D}}^t \neq \emptyset$ indicate that the $\text{IR}$ or  $\text{D}$ modality is available in $X_T$.
 After adapting the model $\textbf{T}$ to the target samples, we then conduct inference on the  target samples.

In Figure~\ref{fig:Framework}, MFAS-DANet consists of three feature extractors $\mathbf{F}_\text{RGB}, \mathbf{F}_\text{IR}$, and $\mathbf{F}_\text{D}$, three classifiers $\mathbf{C}_\text{RGB}, \mathbf{C}_\text{IR}$, and $\mathbf{C}_\text{D}$, and three feature adapter $\mathbf{A}_\text{RGB}, \mathbf{A}_\text{IR}$, and $\mathbf{A}_\text{D}$.
First, to address the concern of missing modalities, we propose an effective cross-modal feature transformation strategy during  the adaptation processes. Next, to address the noisy pseudo labels, we propose a novel prediction-reliability aware pseudo-labeling to determine the final pseudo-labels based on the reliability of prediction. Finally, we propose an effective model-stability aware adaptation technique to mitigate the risk of model degradation during adaptation, while simultaneously enabling the model to learn new attack types and unseen domain information.

\subsection{Cross-modal feature transformation}
\label{sec:feature enhancement}
Given the complementary nature of different modalities, our central idea for the adaptation process is  to use  complementary features transformed from other modalities to compensate for the features of missing modalities or to enhance the features of existing modalities.

First,  we employ the feature extractors $\mathbf{F}_\text{RGB}, \mathbf{F}_\text{IR}$, and $\mathbf{F}_\text{D}$ to extract  multi-modal liveness features $\mathbf{f}^t_\text{mod}$ by
\begin{eqnarray}
\mathbf{f}^t_\text{mod} = \mathbb{I}(\mathbf{x}_\text{mod}^t \neq \emptyset) \cdot \mathbf{F}_\text{mod} (\mathbf{x}^t_\text{mod})
\end{eqnarray}

\noindent
where $\text{mod} \in \{ \text{RGB}, \text{IR}, \text{D}\}$ denotes the different modalities. 
 
Next, as noted in existing FAS method \cite{liu2021face} that RGB and IR images offer rich and complementary texture information, we propose using feature adapters $\mathbf{A}_\text{RGB}$ and $\mathbf{A}_\text{IR}$ to  transfer  the complementary texture features $\hat{\mathbf{f}}_\text{RGB}^s$, $\hat{\mathbf{f}}_\text{IR}^s$, $\hat{\mathbf{f}}_\text{RGB}^t$ and $\hat{\mathbf{f}}_\text{IR}^t$ from $ \mathbf{f}_\text{IR}^s$, $\mathbf{f}_\text{RGB}^s$, $ \mathbf{f}_\text{IR}^t$ and $\mathbf{f}_\text{RGB}^t$ by, 
\begin{eqnarray}
\hat{\mathbf{f}}^s_\text{RGB} = \mathbf{A}_\text{RGB} (\mathbf{f}^s_\text{IR}), \hat{\mathbf{f}}^s_\text{IR} = \mathbf{A}_\text{IR} (\mathbf{f}^s_\text{RGB}), \\
\hat{\mathbf{f}}^t_\text{RGB} = \mathbf{A}_\text{RGB} (\mathbf{f}^t_\text{IR}), \hat{\mathbf{f}}^t_\text{IR} = \mathbf{A}_\text{IR} (\mathbf{f}^t_\text{RGB})
\end{eqnarray} 

\noindent
where $\hat{\mathbf{f}}^s_\text{RGB}$, $\hat{\mathbf{f}}^t_\text{RGB}$, $\hat{\mathbf{f}}^s_\text{IR}$, and $\hat{\mathbf{f}}^t_\text{IR}$ are the transformed RGB and IR features.

To ensure alignment of these transformed features with their specific modalities, we keep other modules fixed and  introduce regularization losses $\mathcal{L}_{{A}_\text{RGB}}$ and  $\mathcal{L}_{{A}_\text{IR}}$  to train  the feature adapters $\mathbf{A}_\text{RGB}$ and $\mathbf{A}_\text{IR}$ by, 
\begin{eqnarray}
\label{eq:loss_RGB} 
    \mathcal{L}_{{A}_\text{RGB}} &=& \sum_{\forall \mathbf{f}_\text{RGB}^t} \mathbb{I}(\mathbf{x}_\text{IR}^t \neq \emptyset) \cdot |1 - \cos(\mathbf{f}_\text{RGB}^t, \hat{\mathbf{f}}_\text{RGB}^t )| + \sum_{\forall \mathbf{f}_\text{RGB}^s}  |1 - \cos(\mathbf{f}_\text{RGB}^s, \hat{\mathbf{f}}_\text{RGB}^s ) |, \\ 
\label{eq:loss_IR} 
    \mathcal{L}_{{A}_\text{IR}} &=& \sum_{\forall \mathbf{f}_\text{IR}^t} \mathbb{I}(\mathbf{x}_\text{IR}^t \neq \emptyset) \cdot |1 - \cos(\mathbf{f}_\text{IR}^t, \hat{\mathbf{f}}_\text{IR}^t )| + \sum_{\forall \mathbf{f}_\text{IR}^s}  |1 - \cos(\mathbf{f}_\text{IR}^s, \hat{\mathbf{f}}_\text{IR}^s ) |, 
\end{eqnarray} 

\noindent
where    cos($\cdot$) denotes the cosine similarity.
In Eq.\eqref{eq:loss_RGB} and \eqref{eq:loss_IR}, we align the features of specific modalities with their transformed complementary features across both the target domain (first term) and the source domains (second term).

Furthermore, as noted in monocular depth estimation methods \cite{li2020estimate,dong2022towards} that RGB and IR modalities contain potential depth information, 
we keep other modules fixed and  introduce a regularization loss $\mathcal{L}_{{A}_\text{D}}$  to train the depth adapter $\mathbf{A}_\text{D}$ for transferring complementary features  
$\hat{\mathbf{f}}_\text{DR}^s$, $\hat{\mathbf{f}}_\text{DI}^s$, $\hat{\mathbf{f}}_\text{DR}^t$ and $\hat{\mathbf{f}}_\text{DI}^t$  from the RGB and IR features $ \mathbf{f}_\text{RGB}^s$, $\mathbf{f}_\text{IR}^s$, $ \mathbf{f}_\text{RGB}^t$ and $\mathbf{f}_\text{IR}^t$ by,   
\begin{eqnarray} 
\label{eq:loss_D} 
    \mathcal{L}_{{A}_\text{D}} &=& \sum_{\forall \mathbf{f}_\text{D}^t} (  
     \mathbb{I}(\mathbf{x}_\text{D}^t \neq \emptyset) \cdot |1 - \cos(\mathbf{f}_\text{D}^t, \hat{\mathbf{f}}_\text{DR}^t ) | +    \nonumber  \mathbb{I}(\mathbf{x}_\text{D}^t \neq \emptyset) \cdot \mathbb{I}(\mathbf{x}_\text{IR}^t \neq \emptyset) \cdot |1 - \cos(\mathbf{f}_\text{D}^t, \hat{\mathbf{f}}_\text{DI}^t ) | ) 
      \nonumber \\
     && +  \sum_{\forall \mathbf{f}_\text{D}^s}  ( 
     |1 - \cos(\mathbf{f}_\text{D}^s, \hat{\mathbf{f}}_\text{DR}^s ) | +  |1 - \cos(\mathbf{f}_\text{D}^s, \hat{\mathbf{f}}_\text{DI}^s ) |), 
\end{eqnarray} 

\noindent
where 
\begin{eqnarray}
\hat{\mathbf{f}}^s_\text{DR} = \mathbf{A}_\text{D} (\mathbf{f}^s_\text{RGB}), \hat{\mathbf{f}}^s_\text{DI} = \mathbf{A}_\text{D} (\mathbf{f}^s_\text{IR}), \\ 
\hat{\mathbf{f}}^t_\text{DR} = \mathbf{A}_\text{D} (\mathbf{f}^t_\text{RGB}), \hat{\mathbf{f}}^t_\text{DI} = \mathbf{A}_\text{D} (\mathbf{f}^t_\text{IR})
\end{eqnarray} 

\noindent
denotes the transformed depth features.

In both the adaptation and inference stages, we fuse different complementary features to either compensate for missing modalities or enhance the features of existing modalities by,  
\begin{normalsize}  
\begin{eqnarray}  
\small
\Tilde{\mathbf{f}}_\text{RGB} &=& \frac{\mathbf{f}^t_\text{RGB} + \mathbb{I}(\mathbf{x}_\text{IR} \neq \emptyset) \cdot \hat{\mathbf{f}}^t_\text{RGB}}{1 + \mathbb{I}(\mathbf{x}_\text{IR} \neq \emptyset)}, \nonumber \\  [5pt] 
\Tilde{\mathbf{f}}_\text{IR} &=&  \frac{\mathbb{I}(\mathbf{x}_\text{IR} \neq \emptyset) \cdot \mathbf{f}^t_\text{IR} + \hat{\mathbf{f}}^t_\text{IR}}{1 + \mathbb{I}(\mathbf{x}_\text{IR} \neq \emptyset)} \nonumber  \\ [5pt]
\label{eq:feature_fusion} 
\Tilde{\mathbf{f}}_\text{D} &=& \frac{
\mathbb{I}(\mathbf{x}^t_\text{D} \neq \emptyset) \cdot \mathbf{f}^t_\text{D}
+ \hat{\mathbf{f}}^t_\text{DR}
+ \mathbb{I}(\mathbf{x}^t_\text{IR} \neq \emptyset) \cdot \hat{\mathbf{f}}^t_\text{DI}
}{
1 + \mathbb{I}(\mathbf{x}^t_\text{D} \neq \emptyset) + \mathbb{I}(\mathbf{x}^t_\text{IR} \neq \emptyset)
}  
\end{eqnarray}
\end{normalsize}

\noindent 
where $\mathbb{I}(\cdot)$ denotes the indicator function.

\subsection{Prediction-reliability aware pseudo-labeling}
\label{sec:Prediction-reliability}

After obtaining the fused features $\Tilde{\mathbf{f}}_\text{RGB}$, $\Tilde{\mathbf{f}}_\text{IR}$, and $\Tilde{\mathbf{f}}_\text{D}$, we propose a novel prediction-reliability aware pseudo-labeling method to generate reliable pseudo-labels for effective model adaptation.

First, we use the classifiers $\mathbf{C}_\text{RGB}$, $\mathbf{C}_\text{IR}$, and $\mathbf{C}_\text{D}$ to obtain  multi-modal prediction scores $p_\text{mod} = \mathbf{C}_\text{mod} (\Tilde{\mathbf{f}}_\text{mod})$, where $\Tilde{\mathbf{f}}_\text{mod}$ is the fused features from Eq.\eqref{eq:feature_fusion} and $\text{mod} \in \{ \text{RGB}, \text{IR}, \text{D}\}$.

By simply averaging these scores, we obtain initial pseudo-labels as follows:
\begin{eqnarray} 
\begin{aligned}
\bar{y} = 
\begin{cases}
    1,  &  if \quad  \frac{1}{3}( \sum_{\forall \text{mod} \in \{\text{RGB}, \text{IR}, \text{D}\}} p_\text{mod})   \geq h;\\ 
    0,  &  if \quad \frac{1}{3}( \sum_{\forall \text{mod} \in \{\text{RGB}, \text{IR}, \text{D}\}} p_\text{mod})   < h,
\end{cases}
\end{aligned}
\label{eq:assign_naive}  
\end{eqnarray}

\noindent 
where $h=0.5$ is the threshold, and $\bar{y}$ is the assigned initial pseudo-label ($\bar{y} = 1$ for the live class and $\bar{y} = 0$ for the spoof class).

However, as low-quality modalities may yield unreliable multi-modal features \cite{lin2024suppress}, the above pseudo-labeling may generate significant noisy labels and potentially degrade model performance.  
To address this, we  propose adopting dropout related  techniques \cite{gal2016dropout} to estimate prediction uncertainty and  obtain more  reliable pseudo-labels. 
In particular, we propose performing $K$ predictions using features with random dropout masking to explore feature reliability, and then calculate the variances of the prediction score  $v_{\text{RGB}}, v_{\text{IR}}$ and $v_{\text{D}}$ to reflect prediction  uncertainty by,  
\begin{equation}
\begin{split}
v_\text{mod}  = \frac{1}{K} \sum_{k=1}^{K} (p^k_\text{mod} - \mu_\text{mod})^2, 
\label{eq:variances} 
\end{split}
\end{equation} 

\noindent 
where, for each modality $\text{mod} \in \{ \text{RGB}, \text{IR}, \text{D}\}$, $p_{\text{mod}}^k$ denotes the multi-modal predictions using the  $k$-th dropouted features $\Tilde{\mathbf{f}}_\text{mod}^k$, and $\mu_{\text{mod}}$  is the mean  of the multi-modal prediction scores from $K$ predictions.

For the $j$-th target sample $X_T^{j}=${\{$\textbf{x}_{\text{RGB}}^{t,j}, \mathbb{I}(\mathbf{x}_\text{IR}^{t,j}  \neq  \emptyset) \cdot \textbf{x}_{\text{IR}}^{t,j}, \mathbb{I}(\mathbf{x}_\text{D}^{t,j}  \neq  \emptyset) \cdot \textbf{x}_{\text{D}}^{t,j}$\}}, we obtain a corresponding prediction variance set $Var^j = \{v_{\text{RGB}}^j , v_{\text{IR}}^j, v_{\text{D}}^j\}$ and then measure the prediction certainty weights $w_{\text{RGB}}^j$, $w_{\text{IR}}^j$, and $w_{\text{D}}^j$ from $Var^j$ by,  
\begin{equation}
w_{\text{mod}}^j = 1 - \frac{v_{\text{mod}}^j - \min(Var^j)}{\max(Var^j) - \min(Var^j)}, 
\label{eq:weights}
\end{equation} 

\noindent 
where {$\text{mod} \in \{ \text{RGB}, \text{IR}, \text{D}\}$}. Note that, a larger variance indicates higher prediction uncertainty and consequently leads to a lower weight during the pseudo-label refinement process.

Finally, we adopt the softmax function $ \psi( \cdot)$ to normalize the certainty weights $w_{\text{RGB}}^j$, $w_{\text{IR}}^j$, and $w_{\text{D}}^j$ to determine the final prediction score $\hat{p}^j$ by, 
\begin{eqnarray} 
 \hat{p}^j = \psi (w_{\text{RGB}}^j) \cdot p_\text{RGB}^j +  \psi( w_{\text{IR}}^j ) \cdot p_\text{IR}^j +  \psi( w_{\text{D}}^j) \cdot p_\text{D}^j.
\label{eq:refined_score}
\end{eqnarray}

Hence, we assign the final pseudo-label $\hat{y}^j$ to the $j$-th  target sample $X_T^{j}$ by, 
\begin{eqnarray} 
\begin{aligned}
\hat{y}^j = 
\begin{cases}
    1,  &  if \quad \hat{p}^j   \geq h;\\ 
    0,  &  if \quad \hat{p}^j  < h.
\end{cases}
\end{aligned}
\label{eq:assign}  
\end{eqnarray} 

\subsection{Model-stability aware adaptation}
\label{sec:Model-reliability adaptation}
Given that pseudo-labels derived from Eq. (\ref{eq:assign}) may deviate from the ground truth, we must mitigate the risk of model degradation during adaptation while still enabling the model to learn new attack types and information from unseen domains. 
Hence, we propose an effective model-stability aware adaptation. 
 
Figure~\ref{fig:Framework} shows the core concept of our idea.
We design a weighted factor $\alpha$ to control the extent of model adaptation to balance adaptation to new data with the maintenance of model stability.
First, we adopt the current model $\textbf{T}$ with the parameters  $(\theta_{\textbf{T}})$ to obtain pseudo-labels for both  the current batch  $X_T^i$ and the subsequent batch $X_T^{i+1}$.
Next, we use the current batch  $X_T^i$ and its corresponding pseudo-labels $\hat{y}$ to calculate the cross-entropy loss $\mathcal{L}_{ce\text{-}X_T^i}^{\bm{\theta}_{ \textbf{T}}} $ by, 

\begin{equation}
  \mathcal{L}^{\bm{\theta}_{ \textbf{T}}}_{ce\text{-}X_T^{i}}   =   \sum_{ \substack{\forall  \Tilde{\mathbf{f}}_\text{mod}, \\ \text{mod} \in \{ \text{RGB}, \text{IR}, \text{D}\} }}  - \hat{y} \log(\mathbf{C}_\text{mod}(\Tilde{\mathbf{f}}_\text{mod})). 
\label{eqn:entropy_avg}
\end{equation}

\noindent
Here, $\mathcal{L}_{ce\text{-}X_T^i}^{\bm{\theta}_{ \textbf{T}}} $ is used to update the current model $\textbf{T}$ for obtaining a temporary model $\textbf{T}_p$ with  the parameters $\theta_{\textbf{T}_p}$.
Furthermore, we evaluate both $\textbf{T}$ and  $\textbf{T}_p$  using  the next batch $X_T^{i+1}$ to  calculate the cross-entropy losses $\mathcal{L}_{ce\text{-}X_T^{i+1}}^{\theta_{\textbf{T}}}  $ and $\mathcal{L}_{ce\text{-}X_T^{i+1}}^{\theta_{\textbf{T}_p}} $.  
Moreover, we determine the weighted factor $\alpha^i$ based on these two losses by,
\begin{equation}
\begin{scriptsize} 
    \alpha^i = \frac{1}{\beta} \log(1 + \exp(\beta ( \mathcal{L}_{ce\text{-}X_T^{i+1}}^{\theta_{\textbf{T}}} - \mathcal{L}_{ce\text{-}X_T^{i+1}}^{\theta_{\textbf{T}_p}}  ))),
\label{eq:adaptation_weight}
\end{scriptsize}
\end{equation} 

\noindent 
where $\beta =500$ is the  scaling factor in the softplus function \cite{dugas2000incorporating}. 
In Eq. \eqref{eq:adaptation_weight}, the weighted factor $\alpha^i$ assesses the model's performance  before and after the update (i.e., $\textbf{T}$ and $\textbf{T}_p$) using the next batch $X_T^{i+1}$. 
Therefore, $\alpha^i$ effectively manages the model adaptation process.
Finally, with $\mathbf{A}_\text{RGB}$, $\mathbf{A}_\text{IR}$, and $\mathbf{A}_\text{D}$ fixed, we use $\alpha^i \cdot \mathcal{L}^{\bm{\theta}_{ \textbf{T}}}_{ce\text{-}X_T^{i}} $ to adapt $\textbf{T}$ to $X_T^{i}$.
The complete optimization problem for learning the parameters of $\textbf{T}$ is described by,
\begin{small}
\begin{eqnarray}
    \bm{\theta}^*_{\textbf{T}} =  \argmin_{\bm{\theta}_{ \textbf{T}} } \alpha^i \cdot \mathcal{L}^{\bm{\theta}_{ \textbf{T}}}_{ce\text{-}X_T^{i}}.
\label{eqn:theta_T}
\end{eqnarray}
\end{small}

After adaptation, we use Equations   \eqref{eq:feature_fusion},   \eqref{eq:variances}, \eqref{eq:weights}, and \eqref{eq:refined_score}    to conduct inference and obtain predictions for the target samples. 
Finally, we adopt the Youden Index Calculation \cite{youden1950index}, as in \cite{wang2022domain, yu2020searching}, to determine the binary classification threshold.

\section{Experiments}
\label{sec:experiment}

\subsection{Experiment setting}
We conduct extensive experiments on the following multi-modal face anti-spoofing databases: (a) CASIA-SURF\cite{shifeng2020casia} (\textbf{S}), (b)  CASIA-CeFA\cite{liu2021casia} (\textbf{C}), (c) WMCA\cite{george2019biometric} (\textbf{W}), and (d) PADISI\cite{Rostami_2021_ICCV} (\textbf{P}). 
To ensure a fair comparison with previous multi-modal FAS method \cite{lin2024suppress}, we adopt the same evaluation metrics: HTER (\%) $\downarrow$ \cite{anjos2011counter} and AUC (\%) $\uparrow$.  

In developing the proposed  MFAS-DANet, we follow previous multi-modal FAS methods \cite{george2021cross, yang2020pipenet, yu2020multi} by adopting three feature extractors $\mathbf{F}_\text{RGB}$, $\mathbf{F}_\text{IR}$, and $\mathbf{F}_\text{D}$, to learn modality-specific features.
We use ResNet-34 as the backbone for  $\mathbf{F}_\text{RGB}$, $\mathbf{F}_\text{IR}$, and $\mathbf{F}_\text{D}$, and use a single linear layer to build each of the classifiers $\mathbf{C}_\text{RGB}$, $\mathbf{C}_\text{IR}$, and $\mathbf{C}_\text{D}$. We evaluate our method on the multi-modal FAS benchmark under both fixed and missing modality  scenarios, as proposed in \cite{lin2024suppress}.
We set $K=10$ for performing $K$  predictions to obtain reliable pseudo-labels.  
For training the FAS models on the source data, we set a constant learning rate of 1$\mathbf{e-}$4 with Adam optimizer up to 10 epochs.
During the adaptation stage, we fix the first three blocks of the feature extractor and update the rest of the FAS model with a batch size of 32 and the learning rate of 1$\mathbf{e-}$6 using the Adam optimizer.

\subsection{Ablation study}

\begin{figure}[t]
    \centering
    \begin{minipage}[t]{0.95\textwidth}
       
        \centering
        \fbox{\includegraphics[width=\linewidth]{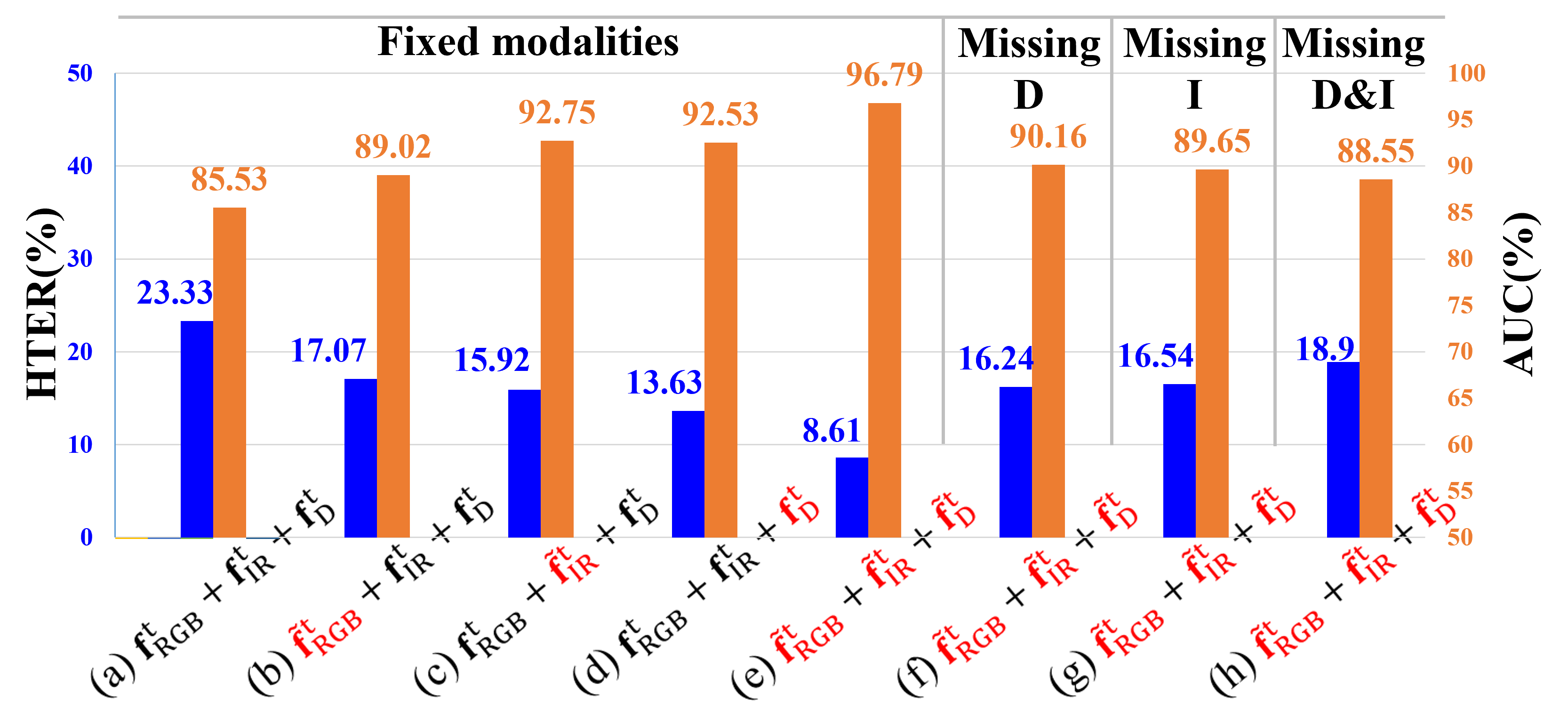}}\\
        \scriptsize  
        \caption{
            Ablation study on \textbf{{[}C,P,S{]}$\rightarrow$W} using different combinations of transformed features.
        }
        \label{fig:ablation_combinations}
    \end{minipage}%
  
\end{figure}

\subsubsection{On different combinations of transformed features} 
Figure~\ref{fig:ablation_combinations} compares different combinations of complementary features in Eq.\eqref{eq:feature_fusion} for adapting FAS models using  pseudo-labels from Eq.\eqref{eq:assign} and the adaptation strategy in Eq.\eqref{eqn:theta_T} for both fixed and missing modality scenarios  proposed in \cite{lin2024suppress}. 
Cases (a)–(e) address the fixed modalities scenario, while (f)–(h) address the missing modality scenario.
In particular, case (a) uses only the original features $\mathbf{f}^t_\text{RGB}+\mathbf{f}^t_\text{IR}+\mathbf{f}^t_\text{D}$ as the baseline. 
Cases (b)-(d) replace  $\mathbf{f}^t_\text{RGB},\mathbf{f}^t_\text{IR},\mathbf{f}^t_\text{D}$ in  (a)  with the fused features $\Tilde{\mathbf{f}}_\text{RGB}, \Tilde{\mathbf{f}}_\text{IR},$ and $\Tilde{\mathbf{f}}_\text{D}$, respectively. 
Cases (e)-(h) use the fused features $\Tilde{\mathbf{f}}_\text{RGB}+ \Tilde{\mathbf{f}}_\text{IR} + \Tilde{\mathbf{f}}_\text{D}$ from Eq.\eqref{eq:feature_fusion} under different missing modality scenarios. 
Comparing (a)–(h) shows that complementary features transformed from other modality features not only enhance original features under the fixed modalities scenario (b)–(e), but also effectively substitute missing modality features to  handle the missing modality scenarios (f)–(h).

\subsubsection{On different pseudo-labeling mechanisms}
Table~\ref{tab:different pseudo-labeling} compares pseudo-labeling mechanisms for adapting FAS models using the fused features from Eq.\eqref{eq:feature_fusion} and the adaptation strategy in Eq.\eqref{eqn:theta_T}.  
First, the class prototype-based approaches \cite{iwasawa2021test, jang2023testtime}, relying on class centers to determine pseudo labels, are insufficient to generate reliable pseudo labels for diverse spoof attacks.
Next, due to the unreliable features extracted from low-quality modalities, particularly depth and infrared \cite{lin2024suppress},  pseudo-labels generated by the score-based method become less informative and unreliable, resulting in poor generalization to unseen target domain attacks.
In contrast, our prediction-reliability aware pseudo-labeling, by estimating prediction uncertainty for reliable labels, outperforms other mechanisms in pseudo-label reliability.

\begin{table}[t]
\centering
\setlength{\tabcolsep}{2pt}
\resizebox{0.9\columnwidth}{!}{
\begin{tabular}{c|cc|cc|cc|cc}
\hline
\textbf{Pseudo-labeling} & \multicolumn{2}{c|}{\textbf{Fixed modalities}} & \multicolumn{2}{c|}{\textbf{Missing D}} & \multicolumn{2}{c|}{\textbf{Missing I}} & \multicolumn{2}{c }{\textbf{Missing D\&I}} \\
\cline{2-9} 
\textbf{mechanisms} & \multicolumn{1}{c|}{HTER} & AUC & \multicolumn{1}{c|}{HTER} & AUC & \multicolumn{1}{c|}{HTER} & AUC & \multicolumn{1}{c|}{HTER} & AUC \\ \hline 
Prototype-based & \multicolumn{1}{c|}{14.45} & 91.30 & \multicolumn{1}{c|}{20.27} & 86.79 & \multicolumn{1}{c|}{20.60} & 86.57 & \multicolumn{1}{c|}{23.13} & 84.26 \\ \hline
Score-based & \multicolumn{1}{c|}{18.18} & 88.12 & \multicolumn{1}{c|}{21.44} & 86.27 & \multicolumn{1}{c|}{21.92} & 84.92 & \multicolumn{1}{c|}{23.82} & 83.66 \\ \hline

\textbf{Ours} & \multicolumn{1}{c|}{\textbf{8.61}} & \textbf{96.76} & \multicolumn{1}{c|}{\textbf{16.24}} & \textbf{90.16} & \multicolumn{1}{c|}{\textbf{16.54}} & \textbf{89.65} & \multicolumn{1}{c|}{\textbf{18.9}} & \textbf{88.55}\\ \hline
\end{tabular}
}
\caption{\small Ablation study on the protocol \textbf{{[}C,P,S{]}$\rightarrow$ W}, using different pseudo-labeling mechanisms.}
\label{tab:different pseudo-labeling}
\end{table}

\begin{table}[t]
\setlength{\tabcolsep}{2pt}
\centering
\resizebox{0.9\columnwidth}{!}{
\begin{tabular}{ c|cc|cc|cc|cc }
\hline
\textbf{Adaptation } & \multicolumn{2}{c|}{\textbf{Fixed modalities}} & \multicolumn{2}{c|}{\textbf{Missing D}} & \multicolumn{2}{c|}{\textbf{Missing I}} & \multicolumn{2}{c }{\textbf{Missing D\&I}} \\
\cline{2-9} 
\textbf{strategies} & \multicolumn{1}{c|}{HTER} & AUC & \multicolumn{1}{c|}{HTER} & AUC & \multicolumn{1}{c|}{HTER} & AUC & \multicolumn{1}{c|}{HTER} & AUC \\ \hline  
$\mathcal{L}^{\bm{\theta}_{ \textbf{T}}}_{ce\text{-}X_T^{i}}$ in \eqref{eqn:entropy_avg} & \multicolumn{1}{c|}{12.07} & 94.02 & \multicolumn{1}{c|}{18.47} & 89.69 & \multicolumn{1}{c|}{19.04} & 88.35 & \multicolumn{1}{c|}{23.46} & 84.10 \\ \hline
$\alpha^i \cdot \mathcal{L}^{\bm{\theta}_{ \textbf{T}}}_{ce\text{-}X_T^{i}}$ in \eqref{eqn:theta_T} & \multicolumn{1}{c|}{\textbf{8.61}} & \textbf{96.76} & \multicolumn{1}{c|}{\textbf{16.24}} & \textbf{90.16} & \multicolumn{1}{c|}{\textbf{16.54}} & \textbf{89.65} & \multicolumn{1}{c|}{\textbf{18.90}} & \textbf{88.55}\\ \hline
\end{tabular}
}
\caption{\small 
Ablation study on the protocol \textbf{{[}C,P,S{]}$\rightarrow$  W}, using different adaptation strategies.
}
\label{tab:different adaptations}
\end{table}

\subsubsection{On different adaptation strategies}
Table~\ref{tab:different adaptations} compares different adaptation strategies for adapting  FAS models using the fused features from Eq.\eqref{eq:feature_fusion} and the pseudo-labels from Eq.\eqref{eq:assign}.
For $\mathcal{L}^{\bm{\theta}_{ \textbf{T}}}_{ce\text{-}X_T^{i}}$, incorrect pseudo-labels tend to degrade the FAS model due to uncontrolled adaptation.
Since pseudo-labels differ from the ground truth, when facing incorrect pseudo-labels,  the FAS model tend to degrade without reasonable control over the extent of model adaptation during the adaptation process.
In contrast, our adaptive weighted factor $\alpha$ balances adaptation and stability, thereby mitigating the risk of model degradation and achieving promising performance.

\begin{figure}[t]  
\centering
\begin{minipage}[b]{0.7\linewidth}
    \centering
    \frame{\includegraphics[width=0.8\linewidth,height=5cm]{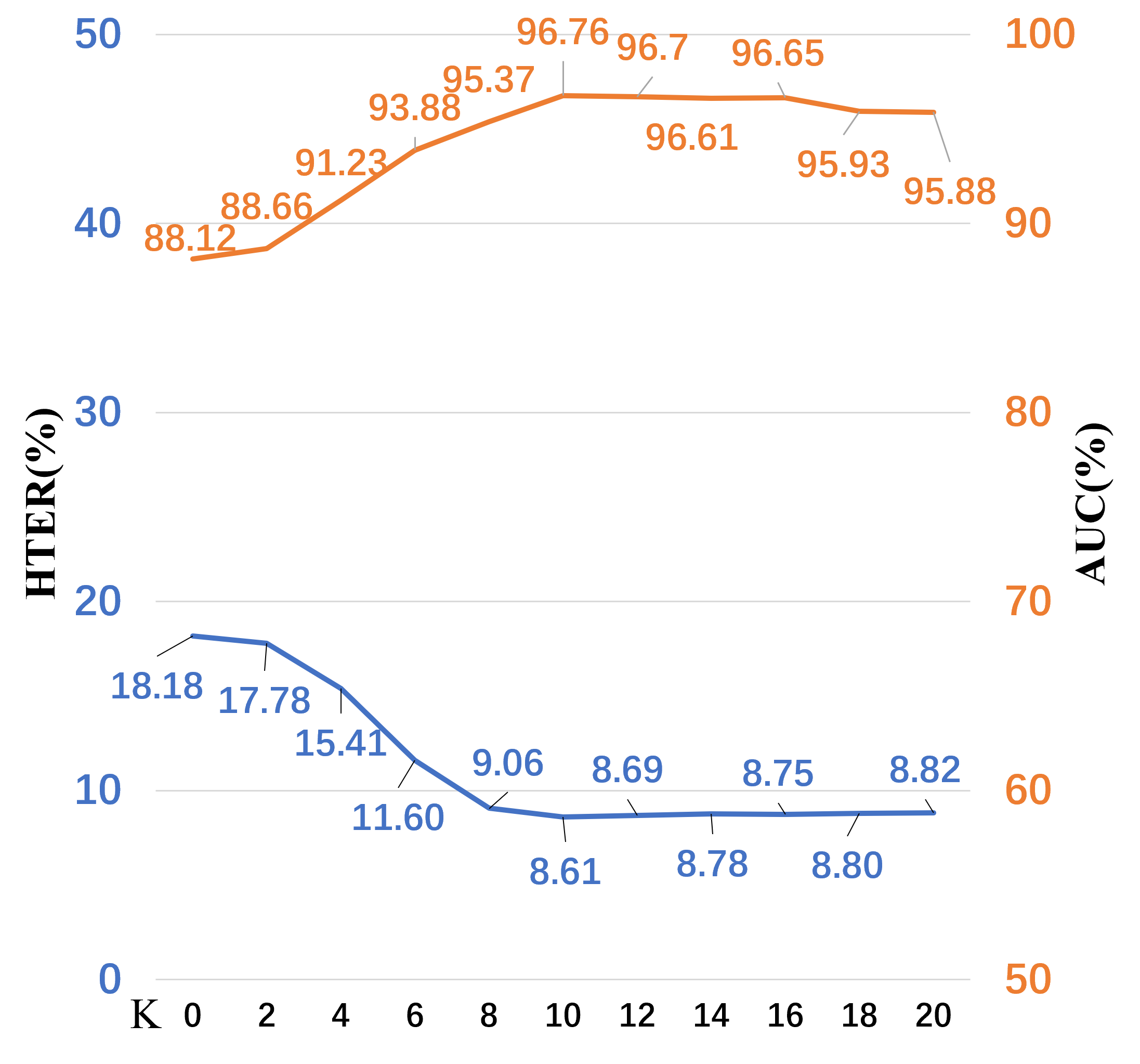}} \\
    (a)
\end{minipage}
\vfill
\begin{minipage}[b]{0.7\linewidth}
    \centering
    \frame{\includegraphics[width=0.7\linewidth,height=6cm]{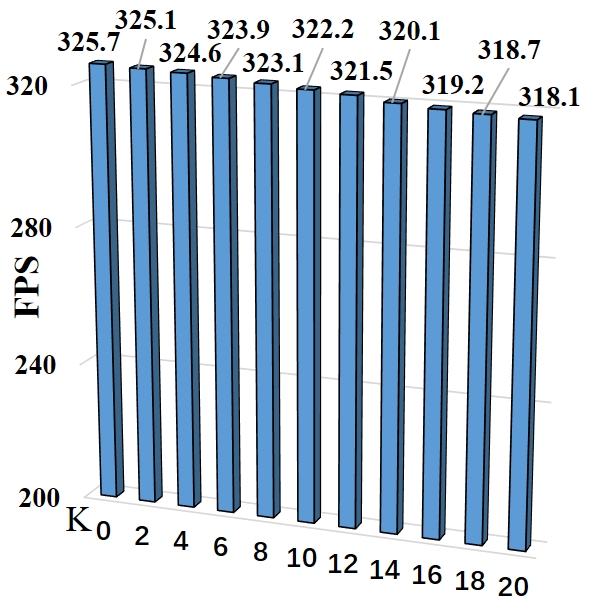}} \\
    (b)
\end{minipage}
\caption{Ablation study on the protocol \textbf{{[}C,P,S{]}$\rightarrow$ W} under fixed modalities scenario, using different $K$ in Eq. (7) during the adaptation and inference stages.}
\label{fig:diff_K}
\end{figure}

\subsubsection{On different $K$ Values}

In Figure~ \ref{fig:diff_K}, we compare using different $K$ in Eq. (7) during the adaptation and inference stages.
Note that, when $K=0$,  the proposed prediction-reliability-aware pseudo-labeling in Eq.(10)  degrades into naive pseudo-labeling in Eq.~(6).
First, in Figure~\ref{fig:diff_K} (a), we find that the performance improves steadily as $K$ increases from 0 to 10 and achieve the best HTER and AUC at $K=10$.
By applying dropout techniques to randomly mask features for exploring feature reliability, we observe that the proposed prediction-reliability-aware pseudo-labeling effectively estimates prediction uncertainty, enabling the acquisition of more reliable pseudo-labels.
Next, we observe that the performance does not improve as $K$ increases from 10 to 20.
This performance plateau indicates that the proposed prediction-reliability-aware pseudo-labeling has reached a state of overcompleteness.
Finally, since we apply dropout only to the fused features from Eq.(5) and only include an additional classifier computation cost of  $K$ times, K increases,  the inference speed in Figure~\ref{fig:diff_K} (b) experiences only a slight decline.

\begin{figure}[t]
    \centering
   
    \begin{minipage}[t]{0.95\textwidth}
        \centering
        \begin{minipage}[t]{0.49\textwidth}
            \centering
            \fbox{\includegraphics[width=\linewidth]{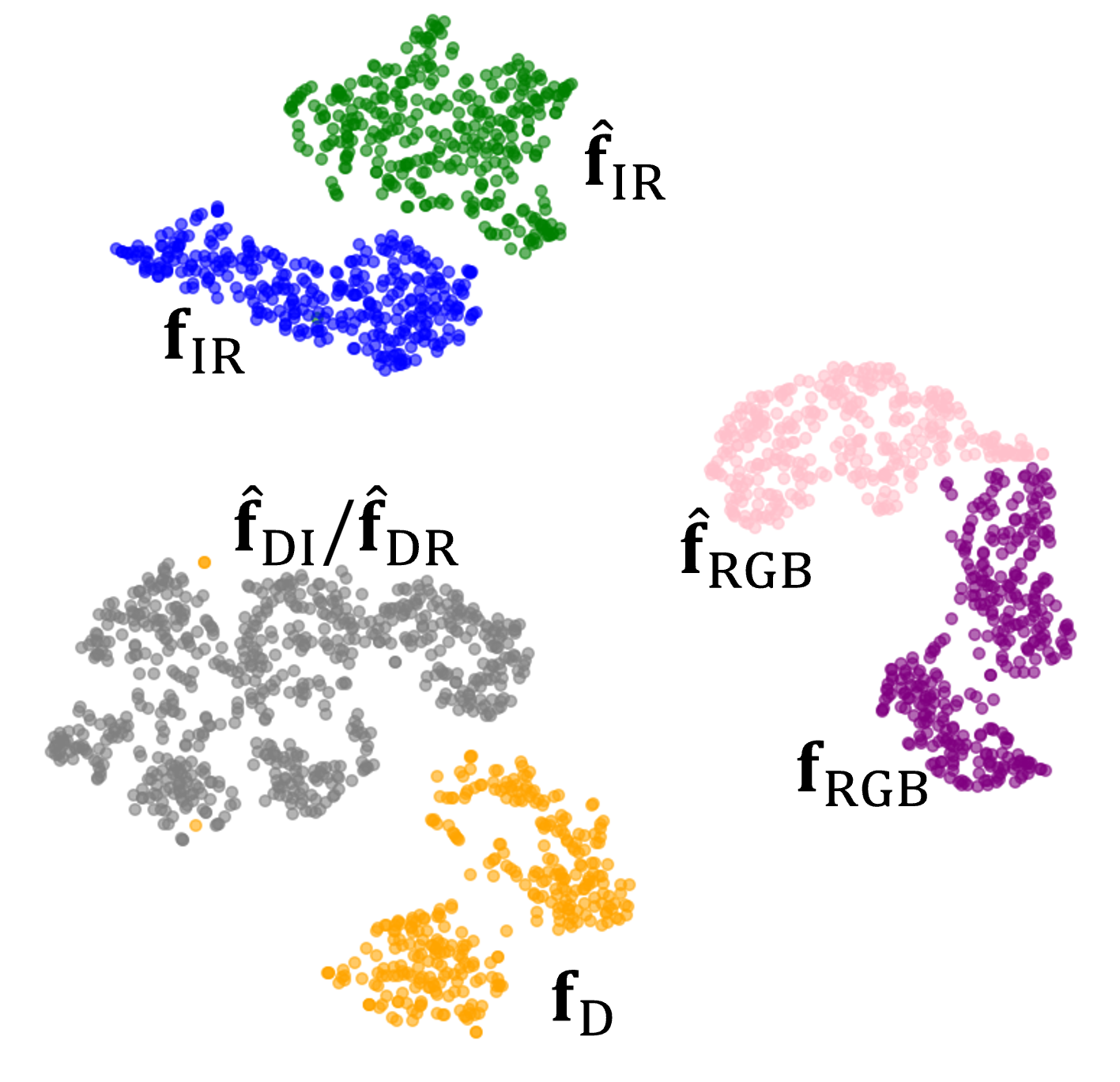}}\\
            \scriptsize (a) 
        \end{minipage}%
        \vfill
        \begin{minipage}[t]{0.49\textwidth}
            \centering
            \fbox{\includegraphics[width=\linewidth]{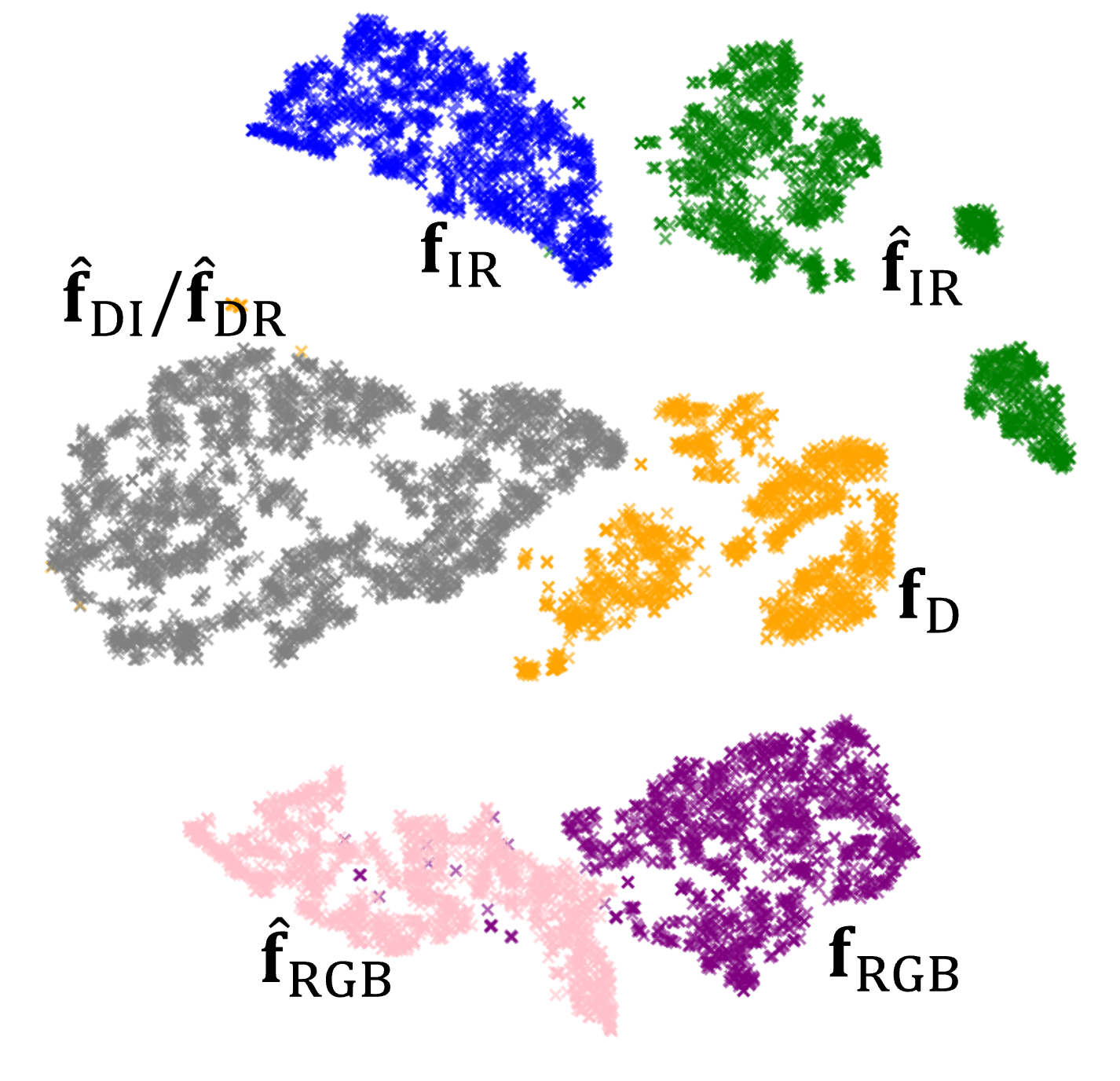}}\\
            \scriptsize (b) 
        \end{minipage}\\[1ex]
        \caption{
            $t$-SNE visualizations  using original and transformed RGB, IR, and depth features, i.e., RGB features (purple:$\mathbf{f}_\text{RGB}$, pink:$\hat{\mathbf{f}}_\text{RGB}$), IR features (blue:$\mathbf{f}_\text{IR}$, green:$\hat{\mathbf{f}}_\text{IR}$), and depth features (yellow:$\mathbf{f}_\text{D}$, gray:$\hat{\mathbf{f}}_\text{DR}$ and $\hat{\mathbf{f}}_\text{DI}$),  for (a) live and (b) spoof classes.
        }
        \label{fig:tsne_visualization}
    \end{minipage}
\end{figure}

\subsubsection{$t$-SNE visualization} 
Figure~\ref{fig:tsne_visualization} shows $t$-SNE\cite{van2008visualizing} visualization of latent liveness features from the adapted model of original vs. transformed complementary features for (a) live and (b) spoof samples. The visualization demonstrates that, for both classes, complementary features from different modalities tightly cluster around the original features of their corresponding modality.

\subsection{Experimental Comparisons} 
To evaluate the effectiveness of the proposed method, we follow the multi-modal FAS protocols in \cite{lin2024suppress} and conduct experiments across various scenarios, including fixed modalities (Table~\ref{tab:Leave-one-out}),  missing modalities (Table~\ref{tab:Missing}), and limited source domains (Table~ \ref{tab:Limited}).  

\begin{table}[!ht]
\centering
\setlength\tabcolsep{1.5 pt}
\scriptsize
\resizebox{\textwidth}{!}{%
\begin{tabular}{ c|c|cc|cc|cc|cc|cc}
\hline
\multirow{2}{*}{\textbf{Method}} 
& \multirow{2}{*}{\textbf{Type}} 
& \multicolumn{2}{c|}{\textbf{{[}C,P,S{]}$\rightarrow$  W}} 
& \multicolumn{2}{c|}{\textbf{{[}C,P,W{]}$\rightarrow$  S}} 
& \multicolumn{2}{c|}{\textbf{{[}C,S,W{]}$\rightarrow$  P}} 
& \multicolumn{2}{c|}{\textbf{{[}P,S,W{]}$\rightarrow$  C}} 
& \multicolumn{2}{c}{\textbf{Average}} \\ \cline{3-12}

& & \multicolumn{1}{c|}{HTER} & \multicolumn{1}{c|}{AUC}   
& \multicolumn{1}{c|}{HTER} & \multicolumn{1}{c|}{AUC}   
& \multicolumn{1}{c|}{HTER} & \multicolumn{1}{c|}{AUC}   
& \multicolumn{1}{c|}{HTER} & \multicolumn{1}{c|}{AUC}   
& \multicolumn{1}{c|}{HTER} & \multicolumn{1}{c}{AUC}    
\\ \hline

CMFL \cite{george2021cross} (\textit{CVPR 21}) & DG  
& \multicolumn{1}{c|}{18.22} & 88.82         
& \multicolumn{1}{c|}{31.20} & 75.66          
& \multicolumn{1}{c|}{26.68} & 80.85         
& \multicolumn{1}{c|}{36.93} & 66.82          
& \multicolumn{1}{c|}{28.26} & 78.04          
\\ \hline

ViT + AMA \cite{yu2024rethinking} (\textit{IJCV 24}) & DG  
& \multicolumn{1}{c|}{17.56} & 88.74          
& \multicolumn{1}{c|}{27.50} & 80.00          
& \multicolumn{1}{c|}{21.18} & 85.51          
& \multicolumn{1}{c|}{47.48} & 55.56          
& \multicolumn{1}{c|}{28.43} & 77.45          
\\ \hline

VP-FAS \cite{yu2024visual} (\textit{TDSC 24}) & DG 
& \multicolumn{1}{c|}{16.26} &  91.22   
& \multicolumn{1}{c|}{24.42} &  81.07  
& \multicolumn{1}{c|}{21.76} &  85.46  
& \multicolumn{1}{c|}{39.35} &  66.55    
& \multicolumn{1}{c|}{25.42} &  81.08   
\\ \hline

MMDG \cite{lin2024suppress} (\textit{CVPR 24}) & DG  
& \multicolumn{1}{c|}{12.79} & 93.83          
& \multicolumn{1}{c|}{15.32} & 92.86          
& \multicolumn{1}{c|}{18.95} & 88.64          
& \multicolumn{1}{c|}{29.93} & 76.52          
& \multicolumn{1}{c|}{19.25} & 87.96          
\\ \hline

PMC \cite{zhang2021progressive} (\textit{TIP 21}) & DA
& \multicolumn{1}{c|}{18.56} & 89.26         
& \multicolumn{1}{c|}{19.55} & 88.77           
& \multicolumn{1}{c|}{25.46} & 78.90       
& \multicolumn{1}{c|}{30.24} & 76.42            
& \multicolumn{1}{c|}{23.45} & 83.34         
\\ \hline 

CCGA-FAS \cite{he2024category} (\textit{TIFS 24})  & DA
& \multicolumn{1}{c|}{15.62} & 91.75 
& \multicolumn{1}{c|}{28.30} & 76.68
& \multicolumn{1}{c|}{26.32} & 81.99
& \multicolumn{1}{c|}{27.81} & 79.29
& \multicolumn{1}{c|}{24.51} & 82.43
\\ \hline

JDOT-FAS \cite{mao2024weighted} (\textit{IJCV 24}) 
& DA  
& \multicolumn{1}{c|}{16.57} & 89.35
& \multicolumn{1}{c|}{18.39} & 89.71
& \multicolumn{1}{c|}{25.15} & 77.51
& \multicolumn{1}{c|}{27.41} & 78.62
& \multicolumn{1}{c|}{21.88} & 83.30 
\\ \hline 

Ours & DA  
& \multicolumn{1}{c|}{\textbf{8.61}} & \textbf{96.76} 
& \multicolumn{1}{c|}{\textbf{9.66}} & \textbf{96.53} 
& \multicolumn{1}{c|}{\textbf{13.01}} & \textbf{93.68}   
& \multicolumn{1}{c|}{\textbf{19.16}} & \textbf{89.13} 
& \multicolumn{1}{c|}{\textbf{12.61}} & \textbf{94.02}
\\ \hline 
\end{tabular}
} 
\caption{Cross-domain testing on the fixed modalities protocols proposed in\cite{lin2024suppress}.} 
\label{tab:Leave-one-out}
\end{table}

\begin{table}[t]
\centering
\scriptsize
\setlength\tabcolsep{1pt}
\begin{minipage}[t]{0.9\textwidth}
\centering
\resizebox{\textwidth}{!}{%
\begin{tabular}{ c|c|cc|cc|cc|cc }
\hline
\multirow{2}{*}{\textbf{Method}} & \multirow{2}{*}{\textbf{Type}} 
& \multicolumn{2}{c|}{\textbf{Missing D}} 
& \multicolumn{2}{c|}{\textbf{Missing I}} 
& \multicolumn{2}{c|}{\textbf{Missing D\&I}} 
& \multicolumn{2}{c}{\textbf{Average}} \\ 
\cline{3-10}
& & \multicolumn{1}{c|}{HTER} & AUC   
& \multicolumn{1}{c|}{HTER} & AUC   
& \multicolumn{1}{c|}{HTER} & AUC   
& \multicolumn{1}{c|}{HTER} & AUC \\ 
\hline
CMFL \cite{george2021cross} (\textit{CVPR 21}) & DG & 31.37 & 74.62 & 30.55 & 75.42 & 31.89 & 74.29 & 31.27 & 74.78 \\ 
ViT + AMA \cite{yu2024rethinking} (\textit{IJCV 24})& DG & 29.25 & 77.70 & 32.30 & 74.06 & 31.48 & 75.82 & 31.01 & 75.86 \\ 
VP-FAS \cite{yu2024visual} (\textit{TDSC 24})& DG & 29.13 & 78.27 & 29.63 & 77.51 & 30.47 & 76.31 & 29.74 & 77.36 \\ 
MMDG \cite{lin2024suppress} (\textit{CVPR 24})& DG & 24.89 & 82.39 & 23.39 & 83.82 & 25.26 & 81.86 & 24.51 & 82.69 \\ 
PMC \cite{zhang2021progressive} (\textit{TIP 21})& DA & 30.79 & 73.59 & 26.09 & 81.43 & 34.88 & 68.22 & 30.59 & 74.41 \\ 
CCGA-FAS \cite{he2024category} (\textit{TIFS 24})& DA & 31.50 & 74.63 & 26.10 & 81.39 & 33.68 & 65.99 & 30.43 & 74.00 \\ 
JDOT-FAS \cite{mao2024weighted} (\textit{IJCV 24})& DA & 30.70 & 73.45 & 31.43 & 73.39 & 34.58 & 68.99 & 32.24 & 71.94 \\ 
\hline
\textbf{Ours} & DA & \textbf{19.51} & \textbf{88.03} & \textbf{19.82} & \textbf{86.28} & \textbf{21.55} & \textbf{83.79} & \textbf{20.29} & \textbf{86.03} \\ 
\hline
\end{tabular}
} 
\caption{
Cross-domain testing on the missing modalities protocols proposed in \cite{lin2024suppress}.
}
\label{tab:Missing}
\end{minipage}
 
\end{table}

\begin{table}[t]
\centering
\scriptsize
\setlength\tabcolsep{1pt} 
\begin{minipage}[t]{0.8\textwidth}
\centering
\resizebox{\textwidth}{!}{
\begin{tabular}{ c|c|cc|cc|cc}
\hline
\multirow{2}{*}{\textbf{Method}} & \multirow{2}{*}{\textbf{Type}} 
& \multicolumn{2}{c|}{\textbf{CW $\rightarrow$ PS}} 
& \multicolumn{2}{c|}{\textbf{PS $\rightarrow$ CW}} 
& \multicolumn{2}{c}{\textbf{Average}} \\ 
\cline{3-8}
& & \multicolumn{1}{c|}{HTER} & AUC 
  & \multicolumn{1}{c|}{HTER} & AUC 
  & \multicolumn{1}{c|}{HTER} & AUC \\ 
\hline
CMFL \cite{george2021cross} (\textit{CVPR 21}) & DG & 31.86 & 72.75 & 39.43 & 63.17 & 35.65 & 67.96 \\ 
ViT + AMA \cite{yu2024rethinking} (\textit{IJCV 24}) & DG & 29.25 & 76.89 & 38.06 & 67.64 & 33.66 & 70.03 \\ 
VP-FAS \cite{yu2024visual} (\textit{TDSC 24}) & DG & 25.90 & 81.79 & 44.37 & 60.83 & 35.14 & 71.31 \\ 
MMDG \cite{lin2024suppress} (\textit{CVPR 24}) & DG & 20.12 & 88.24 & 36.60 & 70.35 & 28.36 & 79.30 \\ 
PMC \cite{zhang2021progressive} (\textit{TIP 21}) & DA & 26.26 & 81.11 & 36.31 & 65.79 & 31.29 & 73.45 \\ 
CCGA-FAS \cite{he2024category} (\textit{TIFS 24})& DA & 32.53 & 68.81 & 34.54 & 70.65 & 33.54 & 69.73 \\ 
JDOT-FAS \cite{mao2024weighted} (\textit{IJCV 24}) & DA & 25.42 & 78.98 & 35.81 & 65.95 & 30.62 & 72.47 \\ 
\hline
\textbf{Ours} & DA & \textbf{15.37} & \textbf{92.88} & \textbf{25.68} & \textbf{80.23} & \textbf{20.53} & \textbf{86.56} \\ 
\hline
\end{tabular}
} 
\caption{\small
Cross-domain testing on the protocols of limited source domains proposed in \cite{lin2024suppress}.
}
\label{tab:Limited}
\end{minipage} 
\end{table}

\subsubsection{Scenario of fixed modalities}
Table~\ref{tab:Leave-one-out} presents cross-domain testing results under the fixed-modality scenario \cite{lin2024suppress}. 
Note that, for single-modal FAS methods CCGA-FAS \cite{he2024category} and JDOT-FAS \cite{mao2024weighted}, we follow \cite{lin2024suppress} to re-implement them by employing a concatenation of multi-modal inputs.
First, under the domain generalization (DG) setting, since different modalities captured by different sensors may exhibit significant domain shifts, the authors in MMDG \cite{lin2024suppress} proposed suppressing unreliable features from low-quality data to outperform CMFL \cite{george2021cross}, ViT + AMA \cite{yu2024rethinking}, and VP-FAS \cite{yu2024visual}.
Similarly, multi-modal FAS under the domain adaptation (DA) setting  faces the same low-quality data issue, which further exacerbates unreliable pseudo-labeling. 
Although  DA methods are able to learn target domain information during the adaptation process, due to the lack of reliable pseudo-labeling,  we see that existing DA methods, including PMC \cite{zhang2021progressive},CCGA-FAS \cite{he2024category}, and JDOT-FAS \cite{mao2024weighted}, still underperform compared to MMDG.
In contrast, our proposed method, by exploring feature reliability to obtain reliable pseudo-labels in Eq.\eqref{eq:assign}  and maintaining model stability during the adaptation process in Eq.\eqref{eqn:theta_T}, outperforms existing single- and multi-modal FAS methods.
 
\subsubsection{Scenario of missing modalities}
In Table~\ref{tab:Missing}, we present cross-domain testing results under the missing-modality scenario \cite{lin2024suppress}. 
Recognizing that multi-modal FAS models may encounter missing modalities during the inference stage,  some methods, such as ViT + AMA \cite{yu2024rethinking}, VP-FAS \cite{yu2024visual}, and MMDG \cite{lin2024suppress}, proposed using zero padding to replace missing modality features, but their performance showed ineffectively for handling this scenario.  
Next, in PMC \cite{zhang2021progressive}, the authors proposed to transfer missing modalities from RGB at the image level.
However, its high computational demands for image modality transferring limit its real-world applicability.
In contrast, our proposed method, by transferring the complementary depth and IR features from the RGB features, effectively handles the missing-modality scenario and outperforms existing methods.  
 
\subsubsection{Scenario of limited source domains}
Table~\ref{tab:Limited} presents cross-domain testing results under the limited source domains scenario\cite{lin2024suppress}. 
Since \textbf{C} contains a wider variety of skin tones (with samples provided in the supplemental material), the protocol \textbf{PS} $\rightarrow$ \textbf{CW} is more challenging than \textbf{CW} $\rightarrow$ \textbf{PS}.
We observe that existing DG methods significantly degrade in performance on this \textbf{PS} $\rightarrow$ \textbf{CW} protocol when encountering unseen skin tones. Similarly, DA methods, limited by the source domains and lacking training data of these skin tones, generate unreliable pseudo-labels and thus lead to ineffective adaptation. In contrast, our method, leveraging reliable pseudo-labels and effectively handling domain shifts, outperforms existing single- and multi-modal FAS methods even with limited source domains.

\section{Conclusion}
\label{sec:Conclusion}

We introduce a novel and practical domain adaptation (DA) framework, MFAS-DANet, for multi-modal face anti-spoofing.
In MFAS-DANet, we first propose learning complementary features from other modalities to substitute missing modality features or enhance existing ones for mitigating the missing modalities issue.  
Next, to tackle the noisy pseudo labels, we leverage prediction uncertainty across  modalities to generate reliable pseudo-labels for effective model adaptation.
Furthermore, we propose using an adaptive weighted factor to stabilize model adaptation and effectively address model degradation.
Our experimental results demonstrate that the proposed MFAS-DANet outperforms existing domain adaptation methods in effectively handling realistic scenarios for multi-modal face anti-spoofing.





\bibliographystyle{elsarticle-num-names} 
\bibliography{reference}

\begin{thebibliography}{48}
\expandafter\ifx\csname natexlab\endcsname\relax\def\natexlab#1{#1}\fi
\providecommand{\url}[1]{\texttt{#1}}
\providecommand{\href}[2]{#2}
\providecommand{\path}[1]{#1}
\providecommand{\DOIprefix}{doi:}
\providecommand{\ArXivprefix}{arXiv:}
\providecommand{\URLprefix}{URL: }
\providecommand{\Pubmedprefix}{pmid:}
\providecommand{\doi}[1]{\href{http://dx.doi.org/#1}{\path{#1}}}
\providecommand{\Pubmed}[1]{\href{pmid:#1}{\path{#1}}}
\providecommand{\bibinfo}[2]{#2}
\ifx\xfnm\relax \def\xfnm[#1]{\unskip,\space#1}\fi
\bibitem[{Huang et~al.(2024)Huang, Chong, Hsu, Hsu, Chiang, Chen, Hsu et~al.}]{huang2024survey}
\bibinfo{author}{P.-K. Huang}, \bibinfo{author}{J.-X. Chong}, \bibinfo{author}{M.-T. Hsu}, \bibinfo{author}{F.-Y. Hsu}, \bibinfo{author}{C.-H. Chiang}, \bibinfo{author}{T.-H. Chen}, \bibinfo{author}{C.-T. Hsu}, et~al.,
\newblock \bibinfo{title}{A survey on deep learning-based face anti-spoofing},
\newblock \bibinfo{journal}{APSIPA Transactions on Signal and Information Processing} \bibinfo{volume}{13} (\bibinfo{year}{2024}).
\bibitem[{Jia et~al.(2020)Jia, Zhang, Shan, and Chen}]{jia2020single}
\bibinfo{author}{Y.~Jia}, \bibinfo{author}{J.~Zhang}, \bibinfo{author}{S.~Shan}, \bibinfo{author}{X.~Chen},
\newblock \bibinfo{title}{Single-side domain generalization for face anti-spoofing},
\newblock in: \bibinfo{booktitle}{Proceedings of the IEEE/CVF Conference on Computer Vision and Pattern Recognition}, \bibinfo{year}{2020}, pp. \bibinfo{pages}{8484--8493}.
\bibitem[{Liao et~al.(2023)Liao, Chen, Liu, Yeh, Hu, and Chen}]{liao2023domain}
\bibinfo{author}{C.-H. Liao}, \bibinfo{author}{W.-C. Chen}, \bibinfo{author}{H.-T. Liu}, \bibinfo{author}{Y.-R. Yeh}, \bibinfo{author}{M.-C. Hu}, \bibinfo{author}{C.-S. Chen},
\newblock \bibinfo{title}{Domain invariant vision transformer learning for face anti-spoofing},
\newblock in: \bibinfo{booktitle}{Proceedings of the IEEE/CVF Winter Conference on Applications of Computer Vision}, \bibinfo{year}{2023}, pp. \bibinfo{pages}{6098--6107}.
\bibitem[{Huang et~al.(2022)Huang, Ni, Ni, and Hsu}]{huang2022learnable}
\bibinfo{author}{P.-K. Huang}, \bibinfo{author}{H.-Y. Ni}, \bibinfo{author}{Y.~Ni}, \bibinfo{author}{C.-T. Hsu},
\newblock \bibinfo{title}{Learnable descriptive convolutional network for face anti-spoofing.},
\newblock in: \bibinfo{booktitle}{BMVC}, \bibinfo{year}{2022}, p. \bibinfo{pages}{239}.
\bibitem[{Huang et~al.(2023)Huang, Chiang, Chong, Chen, Ni, and Hsu}]{huang2023ldcformer}
\bibinfo{author}{P.-K. Huang}, \bibinfo{author}{C.-H. Chiang}, \bibinfo{author}{J.-X. Chong}, \bibinfo{author}{T.-H. Chen}, \bibinfo{author}{H.-Y. Ni}, \bibinfo{author}{C.-T. Hsu},
\newblock \bibinfo{title}{Ldcformer: Incorporating learnable descriptive convolution to vision transformer for face anti-spoofing},
\newblock in: \bibinfo{booktitle}{2023 IEEE International Conference on Image Processing (ICIP)}, \bibinfo{organization}{IEEE}, \bibinfo{year}{2023}, pp. \bibinfo{pages}{121--125}.
\bibitem[{Yu et~al.(2020)Yu, Qin, Li, Wang, Zhao, Lei, and Zhao}]{yu2020multi}
\bibinfo{author}{Z.~Yu}, \bibinfo{author}{Y.~Qin}, \bibinfo{author}{X.~Li}, \bibinfo{author}{Z.~Wang}, \bibinfo{author}{C.~Zhao}, \bibinfo{author}{Z.~Lei}, \bibinfo{author}{G.~Zhao},
\newblock \bibinfo{title}{Multi-modal face anti-spoofing based on central difference networks},
\newblock in: \bibinfo{booktitle}{Proceedings of the IEEE/CVF Conference on Computer Vision and Pattern Recognition Workshops}, \bibinfo{year}{2020}, pp. \bibinfo{pages}{650--651}.
\bibitem[{Lin et~al.(2024)Lin, Wang, Cai, Liu, Fu, Tang, Yu, and Kot}]{lin2024suppress}
\bibinfo{author}{X.~Lin}, \bibinfo{author}{S.~Wang}, \bibinfo{author}{R.~Cai}, \bibinfo{author}{Y.~Liu}, \bibinfo{author}{Y.~Fu}, \bibinfo{author}{W.~Tang}, \bibinfo{author}{Z.~Yu}, \bibinfo{author}{A.~Kot},
\newblock \bibinfo{title}{Suppress and rebalance: Towards generalized multi-modal face anti-spoofing},
\newblock in: \bibinfo{booktitle}{Proceedings of the IEEE/CVF Conference on Computer Vision and Pattern Recognition}, \bibinfo{year}{2024}, pp. \bibinfo{pages}{211--221}.
\bibitem[{Liu et~al.(2023)Liu, Tan, Yu, Zhao, Wan, Liang, Lei, Zhang, Li, and Guo}]{liu2023fm}
\bibinfo{author}{A.~Liu}, \bibinfo{author}{Z.~Tan}, \bibinfo{author}{Z.~Yu}, \bibinfo{author}{C.~Zhao}, \bibinfo{author}{J.~Wan}, \bibinfo{author}{Y.~Liang}, \bibinfo{author}{Z.~Lei}, \bibinfo{author}{D.~Zhang}, \bibinfo{author}{S.~Z. Li}, \bibinfo{author}{G.~Guo},
\newblock \bibinfo{title}{Fm-vit: Flexible modal vision transformers for face anti-spoofing},
\newblock \bibinfo{journal}{IEEE Transactions on Information Forensics and Security} \bibinfo{volume}{18} (\bibinfo{year}{2023}) \bibinfo{pages}{4775--4786}.
\bibitem[{Liu and Liang(2023)}]{liu2023ma}
\bibinfo{author}{A.~Liu}, \bibinfo{author}{Y.~Liang},
\newblock \bibinfo{title}{Ma-vit: Modality-agnostic vision transformers for face anti-spoofing},
\newblock \bibinfo{journal}{International Joint Conference on Artificial Intelligence}  (\bibinfo{year}{2023}).
\bibitem[{Rostami et~al.(2021)Rostami, Spinoulas, Hussein, Mathai, and Abd-Almageed}]{Rostami_2021_ICCV}
\bibinfo{author}{M.~Rostami}, \bibinfo{author}{L.~Spinoulas}, \bibinfo{author}{M.~Hussein}, \bibinfo{author}{J.~Mathai}, \bibinfo{author}{W.~Abd-Almageed},
\newblock \bibinfo{title}{Detection and continual learning of novel face presentation attacks},
\newblock in: \bibinfo{booktitle}{Proceedings of the IEEE/CVF International Conference on Computer Vision (ICCV)}, \bibinfo{year}{2021}, pp. \bibinfo{pages}{14851--14860}.
\bibitem[{George et~al.(2019)George, Mostaani, Geissenbuhler, Nikisins, Anjos, and Marcel}]{george2019biometric}
\bibinfo{author}{A.~George}, \bibinfo{author}{Z.~Mostaani}, \bibinfo{author}{D.~Geissenbuhler}, \bibinfo{author}{O.~Nikisins}, \bibinfo{author}{A.~Anjos}, \bibinfo{author}{S.~Marcel},
\newblock \bibinfo{title}{Biometric face presentation attack detection with multi-channel convolutional neural network},
\newblock \bibinfo{journal}{IEEE transactions on information forensics and security} \bibinfo{volume}{15} (\bibinfo{year}{2019}) \bibinfo{pages}{42--55}.
\bibitem[{Wang et~al.(2021)Wang, Zhang, Bian, Cai, Wang, and Pu}]{wang2021self}
\bibinfo{author}{J.~Wang}, \bibinfo{author}{J.~Zhang}, \bibinfo{author}{Y.~Bian}, \bibinfo{author}{Y.~Cai}, \bibinfo{author}{C.~Wang}, \bibinfo{author}{S.~Pu},
\newblock \bibinfo{title}{Self-domain adaptation for face anti-spoofing},
\newblock in: \bibinfo{booktitle}{Proceedings of the AAAI conference on artificial intelligence}, volume~\bibinfo{volume}{35}, \bibinfo{year}{2021}, pp. \bibinfo{pages}{2746--2754}.
\bibitem[{Yu et~al.(2023)Yu, Liu, Zhao, Cheng, Cheng, and Zhao}]{yu2023flexible}
\bibinfo{author}{Z.~Yu}, \bibinfo{author}{A.~Liu}, \bibinfo{author}{C.~Zhao}, \bibinfo{author}{K.~H. Cheng}, \bibinfo{author}{X.~Cheng}, \bibinfo{author}{G.~Zhao},
\newblock \bibinfo{title}{Flexible-modal face anti-spoofing: A benchmark},
\newblock in: \bibinfo{booktitle}{Proceedings of the IEEE/CVF Conference on Computer Vision and Pattern Recognition}, \bibinfo{year}{2023}, pp. \bibinfo{pages}{6346--6351}.
\bibitem[{Huang et~al.(2021)Huang, Chin, and Hsu}]{huang2021face}
\bibinfo{author}{P.-K. Huang}, \bibinfo{author}{M.-C. Chin}, \bibinfo{author}{C.-T. Hsu},
\newblock \bibinfo{title}{Face anti-spoofing via robust auxiliary estimation and discriminative feature learning},
\newblock in: \bibinfo{booktitle}{Asian Conference on Pattern Recognition}, \bibinfo{organization}{Springer}, \bibinfo{year}{2021}, pp. \bibinfo{pages}{443--458}.
\bibitem[{Liu et~al.(2018)Liu, Jourabloo, and Liu}]{liu2018learning}
\bibinfo{author}{Y.~Liu}, \bibinfo{author}{A.~Jourabloo}, \bibinfo{author}{X.~Liu},
\newblock \bibinfo{title}{Learning deep models for face anti-spoofing: Binary or auxiliary supervision},
\newblock in: \bibinfo{booktitle}{Proceedings of the IEEE conference on computer vision and pattern recognition}, \bibinfo{year}{2018}, pp. \bibinfo{pages}{389--398}.
\bibitem[{Shao et~al.(2019)Shao, Lan, Li, and Yuen}]{shao2019multi}
\bibinfo{author}{R.~Shao}, \bibinfo{author}{X.~Lan}, \bibinfo{author}{J.~Li}, \bibinfo{author}{P.~C. Yuen},
\newblock \bibinfo{title}{Multi-adversarial discriminative deep domain generalization for face presentation attack detection},
\newblock in: \bibinfo{booktitle}{Proceedings of the IEEE/CVF conference on computer vision and pattern recognition}, \bibinfo{year}{2019}, pp. \bibinfo{pages}{10023--10031}.
\bibitem[{Yu et~al.(2020)Yu, Li, Niu, Shi, and Zhao}]{yu2020face}
\bibinfo{author}{Z.~Yu}, \bibinfo{author}{X.~Li}, \bibinfo{author}{X.~Niu}, \bibinfo{author}{J.~Shi}, \bibinfo{author}{G.~Zhao},
\newblock \bibinfo{title}{Face anti-spoofing with human material perception},
\newblock in: \bibinfo{booktitle}{European Conference on Computer Vision}, \bibinfo{organization}{Springer}, \bibinfo{year}{2020}, pp. \bibinfo{pages}{557--575}.
\bibitem[{Zhou et~al.(2023)Zhou, Zhang, Yao, Lu, Yi, Ding, and Ma}]{zhou2023instance}
\bibinfo{author}{Q.~Zhou}, \bibinfo{author}{K.-Y. Zhang}, \bibinfo{author}{T.~Yao}, \bibinfo{author}{X.~Lu}, \bibinfo{author}{R.~Yi}, \bibinfo{author}{S.~Ding}, \bibinfo{author}{L.~Ma},
\newblock \bibinfo{title}{Instance-aware domain generalization for face anti-spoofing},
\newblock in: \bibinfo{booktitle}{Proceedings of the IEEE/CVF Conference on Computer Vision and Pattern Recognition}, \bibinfo{year}{2023}, pp. \bibinfo{pages}{20453--20463}.
\bibitem[{Liu et~al.(2018)Liu, Lan, and Yuen}]{liu2018remote}
\bibinfo{author}{S.-Q. Liu}, \bibinfo{author}{X.~Lan}, \bibinfo{author}{P.~C. Yuen},
\newblock \bibinfo{title}{Remote photoplethysmography correspondence feature for 3d mask face presentation attack detection},
\newblock in: \bibinfo{booktitle}{Proceedings of the European Conference on Computer Vision (ECCV)}, \bibinfo{year}{2018}.
\bibitem[{Liu et~al.(2022)Liu, Lan, and Yuen}]{liu2022learning}
\bibinfo{author}{S.-Q. Liu}, \bibinfo{author}{X.~Lan}, \bibinfo{author}{P.~C. Yuen},
\newblock \bibinfo{title}{Learning temporal similarity of remote photoplethysmography for fast 3d mask face presentation attack detection},
\newblock \bibinfo{journal}{IEEE Transactions on Information Forensics and Security} \bibinfo{volume}{17} (\bibinfo{year}{2022}) \bibinfo{pages}{3195--3210}.
\bibitem[{Huang et~al.(2022)Huang, Sun, Liu, Chu, Xiao, Yuan, Adam, and Yang}]{huang2022adaptive}
\bibinfo{author}{H.-P. Huang}, \bibinfo{author}{D.~Sun}, \bibinfo{author}{Y.~Liu}, \bibinfo{author}{W.-S. Chu}, \bibinfo{author}{T.~Xiao}, \bibinfo{author}{J.~Yuan}, \bibinfo{author}{H.~Adam}, \bibinfo{author}{M.-H. Yang},
\newblock \bibinfo{title}{Adaptive transformers for robust few-shot cross-domain face anti-spoofing},
\newblock in: \bibinfo{booktitle}{European Conference on Computer Vision}, \bibinfo{organization}{Springer}, \bibinfo{year}{2022}, pp. \bibinfo{pages}{37--54}.
\bibitem[{Huang et~al.(2025{\natexlab{a}})Huang, Chong, Hsu, Hsu, and Hsu}]{huang2025channel}
\bibinfo{author}{P.-K. Huang}, \bibinfo{author}{J.-X. Chong}, \bibinfo{author}{M.-T. Hsu}, \bibinfo{author}{F.-Y. Hsu}, \bibinfo{author}{C.-T. Hsu},
\newblock \bibinfo{title}{Channel difference transformer for face anti-spoofing},
\newblock \bibinfo{journal}{Information Sciences} \bibinfo{volume}{702} (\bibinfo{year}{2025}{\natexlab{a}}) \bibinfo{pages}{121904}.
\bibitem[{Huang et~al.(2025{\natexlab{b}})Huang, Chong, Chiang, Chen, Liu, and Hsu}]{huang2025slip}
\bibinfo{author}{P.-K. Huang}, \bibinfo{author}{J.-X. Chong}, \bibinfo{author}{C.-H. Chiang}, \bibinfo{author}{T.-H. Chen}, \bibinfo{author}{T.-L. Liu}, \bibinfo{author}{C.-T. Hsu},
\newblock \bibinfo{title}{Slip: Spoof-aware one-class face anti-spoofing with language image pretraining},
\newblock in: \bibinfo{booktitle}{Proceedings of the AAAI Conference on Artificial Intelligence}, volume~\bibinfo{volume}{39}, \bibinfo{year}{2025}{\natexlab{b}}, pp. \bibinfo{pages}{3697--3706}.
\bibitem[{Huang et~al.(2024)Huang, Chiang, Chen, Chong, Liu, and Hsu}]{huang2024one}
\bibinfo{author}{P.-K. Huang}, \bibinfo{author}{C.-H. Chiang}, \bibinfo{author}{T.-H. Chen}, \bibinfo{author}{J.-X. Chong}, \bibinfo{author}{T.-L. Liu}, \bibinfo{author}{C.-T. Hsu},
\newblock \bibinfo{title}{One-class face anti-spoofing via spoof cue map-guided feature learning},
\newblock in: \bibinfo{booktitle}{Proceedings of the IEEE/CVF Conference on Computer Vision and Pattern Recognition}, \bibinfo{year}{2024}, pp. \bibinfo{pages}{277--286}.
\bibitem[{He et~al.(2024)He, Peng, Cai, Yu, Long, and Lam}]{he2024category}
\bibinfo{author}{Y.~He}, \bibinfo{author}{F.~Peng}, \bibinfo{author}{R.~Cai}, \bibinfo{author}{Z.~Yu}, \bibinfo{author}{M.~Long}, \bibinfo{author}{K.-Y. Lam},
\newblock \bibinfo{title}{Category-conditional gradient alignment for domain adaptive face anti-spoofing},
\newblock \bibinfo{journal}{IEEE Transactions on Information Forensics and Security}  (\bibinfo{year}{2024}).
\bibitem[{Mao et~al.(2024)Mao, Chen, and Li}]{mao2024weighted}
\bibinfo{author}{S.~Mao}, \bibinfo{author}{R.~Chen}, \bibinfo{author}{H.~Li},
\newblock \bibinfo{title}{Weighted joint distribution optimal transport based domain adaptation for cross-scenario face anti-spoofing},
\newblock \bibinfo{journal}{International Journal of Computer Vision}  (\bibinfo{year}{2024}) \bibinfo{pages}{1--21}.
\bibitem[{Zhou et~al.(2022)Zhou, Zhang, Yao, Yi, Sheng, Ding, and Ma}]{zhou2022generative}
\bibinfo{author}{Q.~Zhou}, \bibinfo{author}{K.-Y. Zhang}, \bibinfo{author}{T.~Yao}, \bibinfo{author}{R.~Yi}, \bibinfo{author}{K.~Sheng}, \bibinfo{author}{S.~Ding}, \bibinfo{author}{L.~Ma},
\newblock \bibinfo{title}{Generative domain adaptation for face anti-spoofing},
\newblock in: \bibinfo{booktitle}{European conference on computer vision}, \bibinfo{organization}{Springer}, \bibinfo{year}{2022}, pp. \bibinfo{pages}{335--356}.
\bibitem[{Yue et~al.(2023)Yue, Wang, Zhang, Feng, Han, Ding, and Wang}]{yue2023cyclically}
\bibinfo{author}{H.~Yue}, \bibinfo{author}{K.~Wang}, \bibinfo{author}{G.~Zhang}, \bibinfo{author}{H.~Feng}, \bibinfo{author}{J.~Han}, \bibinfo{author}{E.~Ding}, \bibinfo{author}{J.~Wang},
\newblock \bibinfo{title}{Cyclically disentangled feature translation for face anti-spoofing},
\newblock in: \bibinfo{booktitle}{Proceedings of the AAAI conference on artificial intelligence}, volume~\bibinfo{volume}{37}, \bibinfo{year}{2023}, pp. \bibinfo{pages}{3358--3366}.
\bibitem[{George and Marcel(2021)}]{george2021cross}
\bibinfo{author}{A.~George}, \bibinfo{author}{S.~Marcel},
\newblock \bibinfo{title}{Cross modal focal loss for rgbd face anti-spoofing},
\newblock in: \bibinfo{booktitle}{Proceedings of the IEEE/CVF conference on computer vision and pattern recognition}, \bibinfo{year}{2021}, pp. \bibinfo{pages}{7882--7891}.
\bibitem[{Yu et~al.(2024{\natexlab{a}})Yu, Cai, Cui, Liu, Hu, and Kot}]{yu2024rethinking}
\bibinfo{author}{Z.~Yu}, \bibinfo{author}{R.~Cai}, \bibinfo{author}{Y.~Cui}, \bibinfo{author}{X.~Liu}, \bibinfo{author}{Y.~Hu}, \bibinfo{author}{A.~C. Kot},
\newblock \bibinfo{title}{Rethinking vision transformer and masked autoencoder in multimodal face anti-spoofing},
\newblock \bibinfo{journal}{International Journal of Computer Vision}  (\bibinfo{year}{2024}{\natexlab{a}}) \bibinfo{pages}{1--22}.
\bibitem[{Yu et~al.(2024{\natexlab{b}})Yu, Cai, Cui, Liu, and Chen}]{yu2024visual}
\bibinfo{author}{Z.~Yu}, \bibinfo{author}{R.~Cai}, \bibinfo{author}{Y.~Cui}, \bibinfo{author}{A.~Liu}, \bibinfo{author}{C.~Chen},
\newblock \bibinfo{title}{Visual prompt flexible-modal face anti-spoofing},
\newblock \bibinfo{journal}{IEEE Transactions on Dependable and Secure Computing}  (\bibinfo{year}{2024}{\natexlab{b}}).
\bibitem[{Antil and Dhiman(2024)}]{antil2024mf2shrt}
\bibinfo{author}{A.~Antil}, \bibinfo{author}{C.~Dhiman},
\newblock \bibinfo{title}{Mf2shrt: multimodal feature fusion using shared layered transformer for face anti-spoofing},
\newblock \bibinfo{journal}{ACM Transactions on Multimedia Computing, Communications and Applications} \bibinfo{volume}{20} (\bibinfo{year}{2024}) \bibinfo{pages}{1--21}.
\bibitem[{Zhang et~al.(2021)Zhang, Xu, Zhang, and Ouyang}]{zhang2021progressive}
\bibinfo{author}{W.~Zhang}, \bibinfo{author}{D.~Xu}, \bibinfo{author}{J.~Zhang}, \bibinfo{author}{W.~Ouyang},
\newblock \bibinfo{title}{Progressive modality cooperation for multi-modality domain adaptation},
\newblock \bibinfo{journal}{IEEE Transactions on Image Processing} \bibinfo{volume}{30} (\bibinfo{year}{2021}) \bibinfo{pages}{3293--3306}.
\bibitem[{Liu et~al.(2021)Liu, Tan, Wan, Liang, Lei, Guo, and Li}]{liu2021face}
\bibinfo{author}{A.~Liu}, \bibinfo{author}{Z.~Tan}, \bibinfo{author}{J.~Wan}, \bibinfo{author}{Y.~Liang}, \bibinfo{author}{Z.~Lei}, \bibinfo{author}{G.~Guo}, \bibinfo{author}{S.~Z. Li},
\newblock \bibinfo{title}{Face anti-spoofing via adversarial cross-modality translation},
\newblock \bibinfo{journal}{IEEE Transactions on Information Forensics and Security} \bibinfo{volume}{16} (\bibinfo{year}{2021}) \bibinfo{pages}{2759--2772}.
\bibitem[{Li et~al.(2020)Li, Ding, Wei, Wu, and Cao}]{li2020estimate}
\bibinfo{author}{X.~Li}, \bibinfo{author}{M.~Ding}, \bibinfo{author}{D.~Wei}, \bibinfo{author}{X.~Wu}, \bibinfo{author}{Y.~Cao},
\newblock \bibinfo{title}{Estimate depth information from monocular infrared images based on deep learning},
\newblock in: \bibinfo{booktitle}{2020 IEEE International Conference on Progress in Informatics and Computing (PIC)}, \bibinfo{organization}{IEEE}, \bibinfo{year}{2020}, pp. \bibinfo{pages}{149--153}.
\bibitem[{Dong et~al.(2022)Dong, Garratt, Anavatti, and Abbass}]{dong2022towards}
\bibinfo{author}{X.~Dong}, \bibinfo{author}{M.~A. Garratt}, \bibinfo{author}{S.~G. Anavatti}, \bibinfo{author}{H.~A. Abbass},
\newblock \bibinfo{title}{Towards real-time monocular depth estimation for robotics: A survey},
\newblock \bibinfo{journal}{IEEE Transactions on Intelligent Transportation Systems} \bibinfo{volume}{23} (\bibinfo{year}{2022}) \bibinfo{pages}{16940--16961}.
\bibitem[{Gal and Ghahramani(2016)}]{gal2016dropout}
\bibinfo{author}{Y.~Gal}, \bibinfo{author}{Z.~Ghahramani},
\newblock \bibinfo{title}{Dropout as a bayesian approximation: Representing model uncertainty in deep learning},
\newblock in: \bibinfo{booktitle}{international conference on machine learning}, \bibinfo{organization}{PMLR}, \bibinfo{year}{2016}, pp. \bibinfo{pages}{1050--1059}.
\bibitem[{Dugas et~al.(2000)Dugas, Bengio, B{\'e}lisle, Nadeau, and Garcia}]{dugas2000incorporating}
\bibinfo{author}{C.~Dugas}, \bibinfo{author}{Y.~Bengio}, \bibinfo{author}{F.~B{\'e}lisle}, \bibinfo{author}{C.~Nadeau}, \bibinfo{author}{R.~Garcia},
\newblock \bibinfo{title}{Incorporating second-order functional knowledge for better option pricing},
\newblock \bibinfo{journal}{Advances in neural information processing systems} \bibinfo{volume}{13} (\bibinfo{year}{2000}).
\bibitem[{Youden(1950)}]{youden1950index}
\bibinfo{author}{W.~J. Youden},
\newblock \bibinfo{title}{Index for rating diagnostic tests},
\newblock \bibinfo{journal}{Cancer} \bibinfo{volume}{3} (\bibinfo{year}{1950}) \bibinfo{pages}{32--35}.
\bibitem[{Wang et~al.(2022)Wang, Wang, Yu, Deng, Li, Gao, and Wang}]{wang2022domain}
\bibinfo{author}{Z.~Wang}, \bibinfo{author}{Z.~Wang}, \bibinfo{author}{Z.~Yu}, \bibinfo{author}{W.~Deng}, \bibinfo{author}{J.~Li}, \bibinfo{author}{T.~Gao}, \bibinfo{author}{Z.~Wang},
\newblock \bibinfo{title}{Domain generalization via shuffled style assembly for face anti-spoofing},
\newblock in: \bibinfo{booktitle}{Proceedings of the IEEE/CVF Conference on Computer Vision and Pattern Recognition}, \bibinfo{year}{2022}, pp. \bibinfo{pages}{4123--4133}.
\bibitem[{Yu et~al.(2020)Yu, Zhao, Wang, Qin, Su, Li, Zhou, and Zhao}]{yu2020searching}
\bibinfo{author}{Z.~Yu}, \bibinfo{author}{C.~Zhao}, \bibinfo{author}{Z.~Wang}, \bibinfo{author}{Y.~Qin}, \bibinfo{author}{Z.~Su}, \bibinfo{author}{X.~Li}, \bibinfo{author}{F.~Zhou}, \bibinfo{author}{G.~Zhao},
\newblock \bibinfo{title}{Searching central difference convolutional networks for face anti-spoofing},
\newblock in: \bibinfo{booktitle}{Proceedings of the IEEE/CVF conference on computer vision and pattern recognition}, \bibinfo{year}{2020}, pp. \bibinfo{pages}{5295--5305}.
\bibitem[{Shifeng et~al.(2020)Shifeng, Ajian, Jun, Yanyan, Guodong, Sergio, and Hugo~Jair}]{shifeng2020casia}
\bibinfo{author}{Z.~Shifeng}, \bibinfo{author}{L.~Ajian}, \bibinfo{author}{W.~Jun}, \bibinfo{author}{L.~Yanyan}, \bibinfo{author}{G.~Guodong}, \bibinfo{author}{E.~Sergio}, \bibinfo{author}{E.~Hugo~Jair},
\newblock \bibinfo{title}{Casia-surf: A large-scale multi-modal benchmark for face anti-spoofing},
\newblock \bibinfo{journal}{IEEE Trans Biometric Behav Identity Sci}  (\bibinfo{year}{2020}).
\bibitem[{Liu et~al.(2021)Liu, Tan, Wan, Escalera, Guo, and Li}]{liu2021casia}
\bibinfo{author}{A.~Liu}, \bibinfo{author}{Z.~Tan}, \bibinfo{author}{J.~Wan}, \bibinfo{author}{S.~Escalera}, \bibinfo{author}{G.~Guo}, \bibinfo{author}{S.~Z. Li},
\newblock \bibinfo{title}{Casia-surf cefa: A benchmark for multi-modal cross-ethnicity face anti-spoofing},
\newblock in: \bibinfo{booktitle}{Proceedings of the IEEE/CVF winter conference on applications of computer vision}, \bibinfo{year}{2021}, pp. \bibinfo{pages}{1179--1187}.
\bibitem[{Anjos and Marcel(2011)}]{anjos2011counter}
\bibinfo{author}{A.~Anjos}, \bibinfo{author}{S.~Marcel},
\newblock \bibinfo{title}{Counter-measures to photo attacks in face recognition: a public database and a baseline},
\newblock in: \bibinfo{booktitle}{2011 international joint conference on Biometrics (IJCB)}, \bibinfo{organization}{IEEE}, \bibinfo{year}{2011}, pp. \bibinfo{pages}{1--7}.
\bibitem[{Yang et~al.(2020)Yang, Zhu, Fwu, Ye, You, and Zhu}]{yang2020pipenet}
\bibinfo{author}{Q.~Yang}, \bibinfo{author}{X.~Zhu}, \bibinfo{author}{J.-K. Fwu}, \bibinfo{author}{Y.~Ye}, \bibinfo{author}{G.~You}, \bibinfo{author}{Y.~Zhu},
\newblock \bibinfo{title}{Pipenet: Selective modal pipeline of fusion network for multi-modal face anti-spoofing},
\newblock in: \bibinfo{booktitle}{Proceedings of the IEEE/CVF conference on computer vision and pattern recognition workshops}, \bibinfo{year}{2020}, pp. \bibinfo{pages}{644--645}.
\bibitem[{Iwasawa and Matsuo(2021)}]{iwasawa2021test}
\bibinfo{author}{Y.~Iwasawa}, \bibinfo{author}{Y.~Matsuo},
\newblock \bibinfo{title}{Test-time classifier adjustment module for model-agnostic domain generalization},
\newblock \bibinfo{journal}{Advances in Neural Information Processing Systems} \bibinfo{volume}{34} (\bibinfo{year}{2021}) \bibinfo{pages}{2427--2440}.
\bibitem[{Jang et~al.(2023)Jang, Chung, and Chung}]{jang2023testtime}
\bibinfo{author}{M.~Jang}, \bibinfo{author}{S.-Y. Chung}, \bibinfo{author}{H.~W. Chung},
\newblock \bibinfo{title}{Test-time adaptation via self-training with nearest neighbor information},
\newblock in: \bibinfo{booktitle}{The Eleventh International Conference on Learning Representations}, \bibinfo{year}{2023}. \URLprefix \url{https://openreview.net/forum?id=EzLtB4M1SbM}.
\bibitem[{Van~der Maaten and Hinton(2008)}]{van2008visualizing}
\bibinfo{author}{L.~Van~der Maaten}, \bibinfo{author}{G.~Hinton},
\newblock \bibinfo{title}{Visualizing data using t-sne.},
\newblock \bibinfo{journal}{Journal of machine learning research} \bibinfo{volume}{9} (\bibinfo{year}{2008}).

\end{thebibliography}
\end{document}